\pdfoutput=1
\PassOptionsToPackage{table}{xcolor}
\documentclass[acmtog,nonacm]{acmart}

\acmJournal{TOG}
\acmBooktitle{}
\usepackage{graphicx}
\usepackage{amsmath}
\usepackage{mathtools}
\usepackage{booktabs}
\usepackage{multirow}
\usepackage{threeparttable}
\usepackage{siunitx}
\usepackage{subcaption}
\usepackage{tabularx}
\usepackage[ruled]{algorithm2e}

\SetAlFnt{\small}
\SetAlCapFnt{\small}
\SetAlCapNameFnt{\small}
\SetAlCapHSkip{0pt}
\usepackage{enumitem}
\usepackage[utf8]{inputenc}
\usepackage[T1]{fontenc}
\usepackage{soul}
\usepackage{float}
\usepackage{stfloats}
\usepackage{placeins}
\usepackage{xcolor}
\definecolor{rank1}{HTML}{B7DDB1}
\definecolor{rank2}{HTML}{DBEED7}

\begin{document}

\newcommand{\etal}{\textit{et al.}}
\newcommand{\motion}{\mathbf{m}}
\newcommand{\motionlatent}{\mathbf{Z}}
\newcommand{\targetlatent}{\mathbf{Z}_{T}}
\newcommand{\partnerlatent}{\mathbf{Z}_{P}}
\newcommand{\targetaudio}{\mathbf{a}^{T}}
\newcommand{\partneraudio}{\mathbf{a}^{P}}
\newcommand{\partnergaze}{\mathbf{g}^{P}}
\newcommand{\partnercontext}{\mathbf{c}^{P}}
\newcommand{\partnerenc}{\mathcal{E}_{P}}
\newcommand{\teach}{\mathrm{teach}}
\newcommand{\stud}{\mathrm{stud}}
\newcommand{\behavior}{\mathbf{b}}
\newcommand{\emotion}{\mathbf{y}^{\mathrm{emo}}}
\newcommand{\AEenc}{\mathcal{E}_{\mathrm{AE}}}
\newcommand{\AEdec}{\mathcal{D}_{\mathrm{AE}}}
\newcommand{\prior}{v_{\theta}}
\newcommand{\observer}{\phi}
\newcommand{\emoobs}{\psi}
\newcommand{\Dlatent}{256}
\newcommand{\Davatar}{256}  %
\newcommand{\Klatent}{25}
\newcommand{\Tframes}{100}
\newcommand{\Dmotion}{61}
\newcommand{\Dbehav}{6}
\newcommand{\Demo}{7}
\newcommand{\Dstyle}{128}
\newcommand{\stylevec}{\mathbf{s}}
\definecolor{junglegreen}{rgb}{0.113, 0.639, 0.5}
\newcommand{\mhc}[1]{}
\newcommand{\mhr}[2]{#2}
\newcommand{\mhe}[1]{#1}
\newcommand{\mhcdone}[1]{}
\newif\ifarxivrev
\newif\ifrevblue
\newif\ifinlinefigs
\newcommand{\rev}[1]{\ifarxivrev{\ifrevblue\color{blue}\fi#1}\fi}
\newcommand{\subonly}[1]{\ifarxivrev\else#1\fi}
\newcommand{\MM}[1]{}
\newcommand{\HR}[1]{}
\newcommand{\hr}[1]{\textcolor{cyan}{#1}}
\newcommand{\kt}[1]{\textcolor{purple}{\textbf{[KT:~#1]}}}

\arxivrevtrue
\inlinefigstrue  %

\title{STEER: Steerable Dyadic Head Avatars}

\author{Kartik Teotia}
\email{kteotia@mpi-inf.mpg.de}
\orcid{0009-0007-6985-7159}
\affiliation{%
  \institution{Max Planck Institute for Informatics and Saarland Informatics Campus}
  \country{Germany}
}

\author{Helge Rhodin}
\email{hrhodin@mpi-inf.mpg.de}
\orcid{0000-0003-2692-0801}
\affiliation{%
  \institution{Max Planck Institute for Informatics}
  \country{Germany}
}

\author{Hyeongwoo Kim}
\email{hkim@mpi-inf.mpg.de}
\orcid{0000-0002-9685-2579}
\affiliation{%
  \institution{Max Planck Institute for Informatics}
  \country{Germany}
}

\author{Marc Habermann}
\email{mhaberma@mpi-inf.mpg.de}
\orcid{0000-0003-3899-7515}
\affiliation{%
  \institution{Max Planck Institute for Informatics and Saarland Informatics Campus}
  \country{Germany}
}

\author{Christian Theobalt}
\email{theobalt@mpi-inf.mpg.de}
\orcid{0000-0001-6104-6625}
\affiliation{%
  \institution{Max Planck Institute for Informatics and Saarland Informatics Campus}
  \country{Germany}
}

\begin{abstract}
    Facial movement and expression are central to face-to-face communication, conveying turn-taking, attention, agreement, and engagement alongside speech. While speech-driven facial animation has made strong progress in lip synchronization and audio-conditioned motion generation, most methods treat conversational behavior as an emergent byproduct of audio, or expose only coarse sequence-level affect control. As a result, key non-verbal channels such as gaze contact and aversion, rhythmic head motion, and emotion remain difficult to explicitly control.

We present STEER, a controllable 3D dyadic motion prior for reactive conversational head avatars. STEER factorizes conversational behavior into explicit controls for gaze, head rhythm, and emotion, allowing users to steer how an avatar listens, reacts, and engages with a conversation partner. Since temporally aligned annotations for these behaviors are not available in public dyadic corpora, we introduce a tracking and annotation pipeline that recovers behavioral pseudo-labels from in-the-wild dyadic video. A causal flow-matching transformer then learns partner-aware target motion conditioned on audio, partner motion, emotion and the proposed behavioral controls.

We further embed STEER in a photorealistic avatar pipeline by extending a Universal Gaussian Head-Avatar Prior with a learned mapping from tracked parametric motion into its avatar-driving space. This enables controllable animation of high-fidelity Gaussian head avatars without re-training the underlying avatar model. STEER outperforms recent dyadic motion baselines on motion quality, dynamics, and diversity, remains competitive on partner coupling, and enables gaze, head-rhythm, and emotion edits together with an interactive live deployment.
\rev{We make our code and dataset annotations available at our \href{https://github.com/Kartik-Teotia/STEER}{webpage}.}

\end{abstract}


\begin{teaserfigure}
\centering
\includegraphics[width=0.9\textwidth]{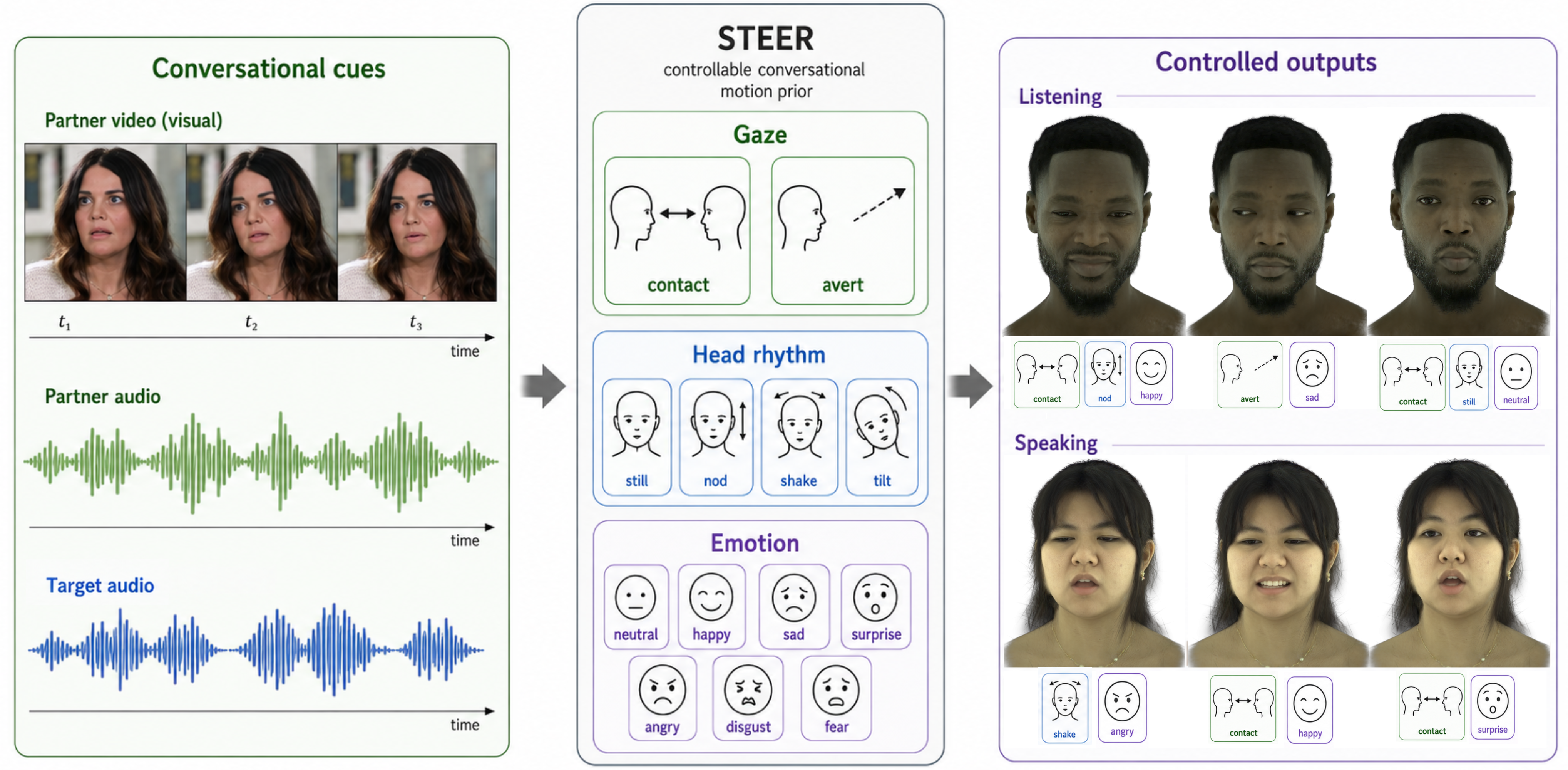}
\caption{STEER takes partner visual/audio cues and the target’s own audio, then generates target motion with explicit semantic controls over gaze, head rhythm, and emotion. The generated motion drives a photoreal Gaussian head avatar, enabling diverse controllable behavior for both listening and speaking roles.}
\label{fig:teaser}
\end{teaserfigure}

\vspace{-0.5em}
 \maketitle

\section{Introduction} \label{sec:intro}
Photoreal conversational head avatars require both responsive dyadic motion and fine-grained semantic control over how the target responds. To the best of our knowledge, no existing system jointly provides partner-aware dyadic generation, explicit temporally local gaze/head-rhythm control, categorical emotion control, and photoreal Gaussian-avatar rendering.
\par
Recent 3D dyadic head motion methods~\cite{peng2025dualtalk,chu2025unils,chatziagapi2025avflow,agrawal2025seamless} have shown that plausible listener and unified speak-listen behavior can be generated from dyadic conversational signals.
However, their non-verbal channel is either shaped implicitly by the conversational input --- AV-Flow~\cite{chatziagapi2025avflow}, for example, produces emergent backchannels with no exposed control and is trained as an identity-specific prior --- or, where control is exposed at all, only at a global level: Seamless Interaction~\cite{agrawal2025seamless} offers controllable variants over global emotion and expressivity, but no factorized interface over non-verbal axes such as gaze contact/aversion or head-rhythm state.
\par
We present STEER, a controllable dyadic motion prior for conversational head avatars.

\par \noindent\textbf{Annotation pipeline.}
Existing public dyadic video corpora do not provide the temporally aligned behavioral annotations needed for explicit control. We therefore introduce an annotation pipeline for in-the-wild dyadic conversations (Sec.~\ref{sec:data}) that combines face and head-pose tracking~\cite{retsinas2024smirk,hempel20226drepnet}, kinematic decomposition~\cite{pan2024s3}, and facial-expression analysis~\cite{chang2024libreface} to recover gaze, head-rhythm, and emotion labels. These pseudo-labels provide the supervision needed to train STEER with explicit control over how an avatar looks, reacts, and expresses affect.

\par \noindent\textbf{Generative prior.}
A causal flow-matching transformer consumes these labels jointly with the target's own audio, partner audio, and partner motion, generating partner-aware target motion while exposing the labels as explicit user controls at inference.
\par \noindent\textbf{Photoreal avatar bridge.}
A separate challenge is realizing this conversational prior in a photoreal avatar pipeline. Speech-driven Gaussian head-avatar models such as UniGAHA~\cite{teotia2025unigaha}, GaussianSpeech~\cite{aneja2025gaussianspeech}, and AudioRTA~\cite{lee2025audiorta} achieve high-fidelity rendering, but they target monadic speech and are not equipped with a controllable conversational motion prior.
Rather than reusing UniGAHA~\cite{teotia2025unigaha} as a black-box renderer, we \emph{extend} its Universal Head-Avatar Prior (UHAP) with a translator that maps our tracked FLAME~\cite{li2017flame} motion space into UHAP's driving latent and adds an explicit gaze-conditioning path that the original UHAP encoder does not expose, allowing our controllable conversational prior to drive a high-fidelity Gaussian avatar without re-training the underlying avatar model.
\par
Our contributions are:
\begin{itemize}
    \item STEER, a dyadic conversational motion prior that exposes interpretable control channels over gaze (contact, avert), head-rhythm (still, nod, shake, tilt), and a seven-class emotion axis, making gaze and head-rhythm directly controllable instead of leaving them implicit in the audio-conditioned prior.
    \item A \emph{tracking and annotation pipeline} for in-the-wild dyadic video that produces temporally aligned labels over gaze, head-rhythm, and emotion, enabling supervised conversational behavior modeling.
    \item An \emph{extension} of a universal photoreal Gaussian head-avatar prior with a learned mapping from tracked FLAME motion to its driving space, embedding our controllable conversational prior in a high-fidelity rendering pipeline without retraining the underlying avatar model.
\end{itemize}
Together, these components make STEER both a stronger dyadic motion prior and a controllable photoreal avatar system. Quantitatively, STEER outperforms recent dyadic motion baselines on motion quality, dynamics, and diversity metrics, while remaining competitive on partner coupling. Qualitatively, Fig.~\ref{fig:teaser} highlights the central capability of STEER: the same partner-aware conversational input can produce diverse, photorealistic target behaviors by explicitly editing gaze, head rhythm, and emotion for both listening and speaking avatars. We further demonstrate an interactive live deployment in the supplemental video, where a user can converse with the avatar and steer its behavior through text or multimodal audio-visual input at $25$\,fps on a single GPU.

\section{Related Work} \label{sec:related_work}
Dyadic head avatars sit at the intersection of conversational motion modeling, semantic control, and photoreal rendering. Existing methods typically address only part of this problem: they may react to a partner, expose some form of control, or render high-fidelity avatars, but rarely combine all three. We review prior work along these axes.
\subsection{Speech-Driven 3D Facial Animation} \label{sec:rw_3dfacial}
A long line of work generates 3D facial animation from audio by mapping acoustic features into the parameter space of a 3D Morphable Model such as FLAME~\cite{li2017flame}: autoregressive and codebook priors~\cite{richard2021meshtalk,fan2022faceformer,xing2023codetalker}, diffusion samplers~\cite{stan2023facediffuser,sun2024diffposetalk,zhao2024media2face,aneja2024facetalk}, and emotion-conditioned variants~\cite{danecek2023emote,peng2023emotalk}. More recently, photoreal speech-driven Gaussian head-avatar models couple a parametric expression code with a 3D Gaussian decoder for direct rendering: \emph{UniGAHA}~\cite{teotia2025unigaha} learns a cross-identity Gaussian decoder driven by a compact expression latent, but is trained on read-sentence audio-visual data and exposes no semantic-control surface over the non-verbal channel; \emph{GaussianSpeech}~\cite{aneja2025gaussianspeech} learns a personalized 3D Gaussian avatar driven by audio with no exposed control; and \emph{AudioRTA}~\cite{lee2025audiorta} drives a similar avatar from audio without head pose or controllable non-verbal axes. These methods target a single active speaker and do not model the asymmetric dynamics of dyadic listening, nor expose a factorized semantic-control interface over non-verbal channels.
\subsection{Listener and Dyadic Conversational Motion Generation} \label{sec:rw_dyadic}
Listener generation has evolved from non-deterministic mappings of speaker audio and motion to listener face sequences~\cite{ng2022learningtolisten,song2023elp,liu2024customlistener,siniukov2025ditailistener,wang2025hybridlhg,lai2025llmlhg} into unified dyadic frameworks that jointly model speaking and listening. \emph{INFP}~\cite{zhu2025infp} dynamically alternates speak/listen states from dyadic audio. \emph{DualTalk}~\cite{peng2025dualtalk} jointly models both roles across multi-round dialogue using partner audio and motion and regresses FLAME blendshapes for the listener. \emph{UniLS}~\cite{chu2025unils} is a two-stage end-to-end dual-track audio-driven avatar. \emph{DyStream}~\cite{chen2025dystream} streams dyadic talking-head generation via a flow-matching autoregressive model for low-latency interactive use; its conditioning is purely audio and it does not expose semantic-control axes. \emph{AV-Flow}~\cite{chatziagapi2025avflow} extends a text-to-audio-visual flow-matching backbone to a dyadic always-on mode that listens to a user's audio-visual stream and emits emergent backchannels, but exposes no explicit control interface over emotion or non-verbal behavior. The recently released \emph{Seamless Interaction}~\cite{agrawal2025seamless} pairs a $4{,}000$-hour dyadic audiovisual corpus with controllable motion models that adapt global emotion and expressivity levels. Yet these controls remain global: they modulate the overall emotion or expressivity of the generated motion, but do not provide time-varying control over gaze contact/aversion or head-rhythm states.
Thus, existing dyadic methods learn when and how a listener reacts, but leave the semantic form of that reaction largely implicit. STEER instead exposes gaze, head rhythm, and emotion as factorized controls, allowing the same conversational context to produce different directed reactions.
\subsection{Controllable Conversational Semantics} \label{sec:rw_control}
Existing control interfaces for conversational face motion fall along three lines: clip-level emotion or expressivity labels~\cite{song2023elp,danecek2023emote,peng2023emotalk,agrawal2025seamless}, free-form text guidance~\cite{liu2024customlistener,siniukov2025ditailistener}, and pose--expression factorization~\cite{wang2025hybridlhg}. Closest to our own decomposition, the animator-centric \emph{S3} system of Pan~\etal~\cite{pan2024s3} splits conversational head-and-eye motion into audio-driven rhythmic head motion, script-driven gestures, and scene-refined gaze trajectories as an authoring abstraction. In the broader body-motion literature, diffusion-based priors conditioned on text, action labels, or style tokens~\cite{tevet2023mdm,guo2024momask} show that coarse semantic control can be layered on a strong generative backbone without sacrificing motion quality. STEER takes up the gaze-versus-rhythm decomposition of \emph{S3} and promotes it from an authoring abstraction into a learned, per-token generative control interface, extending the control surface with a window-level seven-class emotion axis within a single dyadic framework.
\par \noindent\textbf{Summary.}
Across these lines, prior systems miss at least one piece STEER combines: dyadic conditioning, explicit behavioral controls, or photoreal Gaussian-avatar rendering. Speech-driven monadic methods provide a strong audio-to-motion backbone but condition on no partner stream and at most a clip-level emotion control; dyadic methods condition on the partner but do not expose explicit gaze/head-rhythm controls or a reusable bridge to a photoreal Gaussian avatar; photoreal Gaussian-avatar pipelines render high-fidelity output but assume a clean offline driving signal. STEER builds the controllable dyadic prior and the avatar-bridge translator together, so the same model that emits steerable target motion also drives a photoreal renderer at interactive latencies.

\section{Method} \label{sec:method}
\begin{figure*}[t]
\centering
\includegraphics[width=\textwidth]{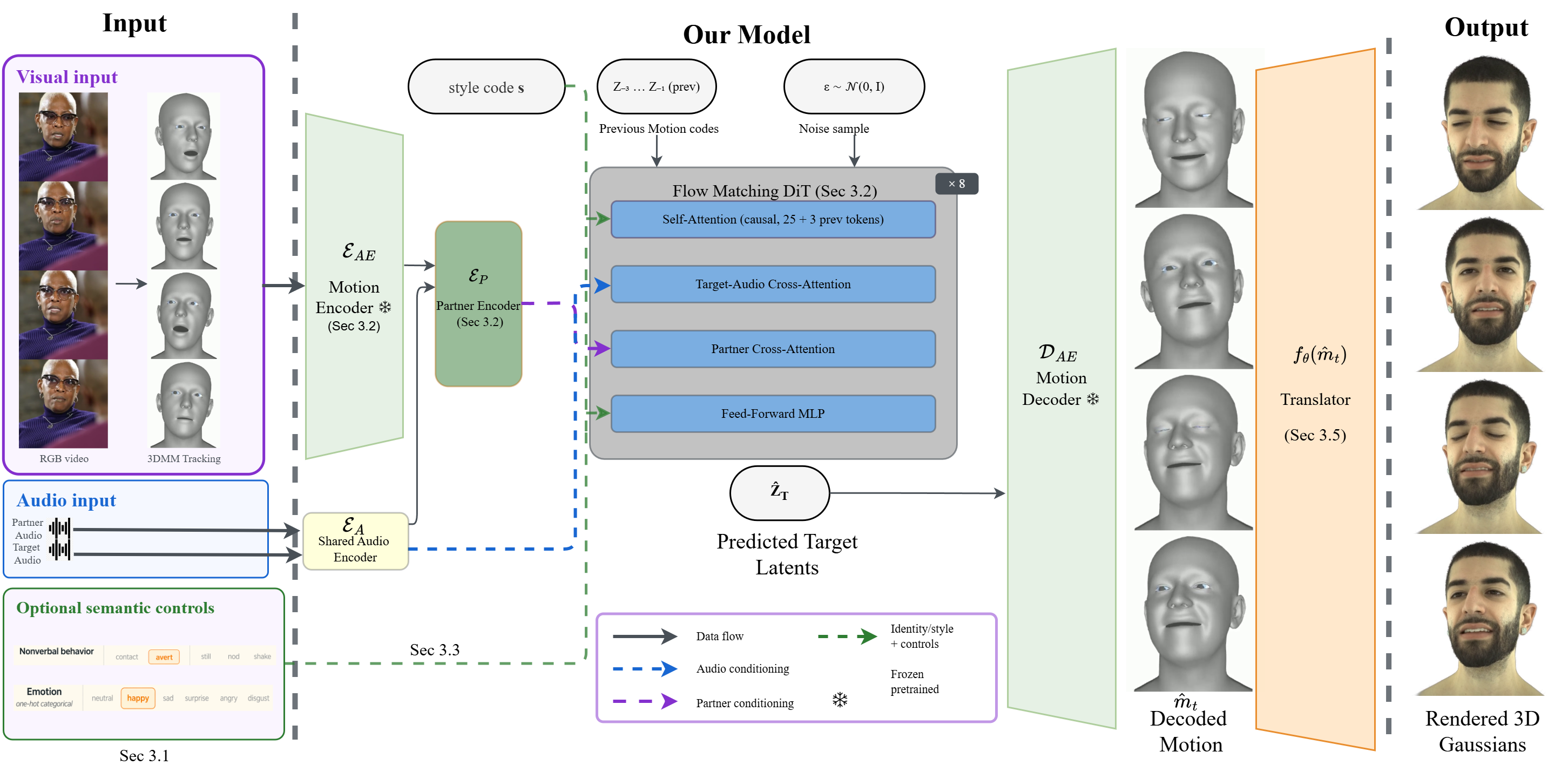}
\caption{
Overview of STEER. A causal flow-matching DiT (center) generates the target's motion as a sequence of $K{=}\Klatent$ latent tokens, conditioned on the target's own audio and a partner context sequence (left) produced by the Partner Encoder over partner audio and partner motion, three semantic control pathways --- style, behavior, emotion (top) --- and a clean prefix of $n_{\mathrm{prev}}{=}3$ previously committed latents for streaming continuity. The decoded motion drives a learned per-frame translator $f_{\theta}$ (right) into a frozen photoreal Gaussian head-avatar prior. Architecture: Sec.~\ref{sec:motion_repr} (motion AE + flow-matching prior), Sec.~\ref{sec:semantic_controls} (semantic controls), Sec.~\ref{sec:train_distill} (streaming inference), Sec.~\ref{sec:translator} (avatar driving).}%
\label{fig:overview}
\end{figure*}
STEER takes streaming partner audio, partner motion, and the target's own audio and produces controllable target motion at $25$\,fps (Fig.~\ref{fig:overview}). We want this motion to react to the partner and to be steerable along semantically meaningful axes --- per-token gaze, per-token head-rhythm, and a window-level emotion --- but existing dyadic corpora lack the supervision such controls require, so we first build an annotation pipeline that derives these labels from in-the-wild dyadic video (Sec.~\ref{sec:data}).
Given the labeled data, we learn a causal flow-matching prior\footnote{By \emph{causal} we mean that any token at time $t$ depends only on inputs up to and including time $t$ --- never on future inputs.} over a motion latent, trained in two phases so the semantic controls do not pre-empt the audio-and-partner backbone (Sec.~\ref{sec:motion_repr}--\ref{sec:train_distill}).
The teacher's iterative inference would prevent real-time deployment, so we distil it into a $2$-step student that preserves the control surface at live rates (Sec.~\ref{sec:distill}).
Finally, the target's motion is more useful when it can drive a photoreal avatar, so we extend UniGAHA's Universal Head-Avatar Prior (UHAP) with a translator that maps tracked FLAME motion into its driving space and adds explicit gaze conditioning (Sec.~\ref{sec:translator}).
\subsection{Dyadic Tracking and Behavior Annotation} \label{sec:data}
Training STEER's controls requires temporally aligned pseudo-labels for gaze and head-rhythm together with a window-level emotion label --- none of which are available in public dyadic corpora. We therefore process each RealTalk~\cite{geng2023affective} shot into a $\Dmotion$-D dynamic FLAME state, extract per-frame audio, and run LibreFace~\cite{chang2024libreface} for two per-frame outputs: $12$-D action-unit intensities and a $7$-class facial-expression prediction (\emph{neutral}, \emph{happy}, \emph{sad}, \emph{angry}, \emph{surprise}, \emph{disgust}, \emph{fear}). We filter unreliable windows, derive per-frame gaze and head-rhythm pseudo-labels and pool them to the latent-token rate downstream (Sec.~\ref{sec:behavior_ctrl}), and aggregate LibreFace's per-frame emotion predictions into a window-level label by smoothing the per-frame argmax over a $7$-frame window, computing a histogram over the $7$ classes, and dropping windows whose dominant class has confidence below $0.45$ from the emotion supervision. Tab.~\ref{tab:tracking_pipeline} summarizes the per-shot extractors and the conditioning or supervision pathway each one feeds.
\par \noindent\textbf{Per-frame motion vector.}
The per-frame state $\motion_t$ stacks SMIRK~\cite{retsinas2024smirk} expression and jaw with neck and eye estimates from external trackers and SMIRK's eyelid openings,
\begin{equation} \label{eq:motion_vec}
\motion_{t} = [\,\mathbf{e}_{t};\ \mathbf{j}_{t};\ \mathbf{n}_{t};\ \mathbf{r}_{t};\ \boldsymbol{\ell}_{t}\,] \in \mathbb{R}^{\Dmotion},
\end{equation}
where $\mathbf{e}_{t} \in \mathbb{R}^{50}$ are the FLAME expression coefficients, $\mathbf{j}_{t}, \mathbf{n}_{t}, \mathbf{r}_{t} \in \mathbb{R}^{3}$ are the jaw, neck, and eye axis-angle rotations with left and right eyes averaged into a single $\mathbf{r}_{t}$, and $\boldsymbol{\ell}_{t} \in \mathbb{R}^{2}$ holds the left and right eyelid openings. %
\begin{table}[h]
\footnotesize
\centering
\caption{Tracking and annotation pipeline. Each per-shot signal is extracted by an off-the-shelf model and consumed by a downstream conditioning or supervision pathway.}
\label{tab:tracking_pipeline}
\setlength{\tabcolsep}{3pt}
\resizebox{\linewidth}{!}{%
\begin{tabular}{@{}llll@{}}
\toprule
{Signal} & {Extractor} & {Output} & {Used for} \\
\midrule
shots & TransNetV2~\cite{soucek2020transnetv2} & per-shot crops & tracking input \\
exp.\ / jaw / eyelid & SMIRK~\cite{retsinas2024smirk} & $\mathbf{e},\mathbf{j},\boldsymbol{\ell}$ & $\motion_t$ \\
neck pose & 6DRepNet~\cite{hempel20226drepnet} & $\mathbf{n}$ & $\motion_t$ \\
eye pose & MediaPipe~\cite{lugaresi2019mediapipe}\,/\,\cite{yan2023mpblendshapes_flame} & $\mathbf{r}$ & $\motion_t$ \\
per-speaker audio & VisualVoice~\cite{gao2021visualvoice} & $16$\,kHz wave & audio cross-attn \\
AU / coarse emotion & LibreFace~\cite{chang2024libreface} & per-frame, $\Demo$-class & emotion control \\
clip filter & Qwen2-VL~\cite{wang2024qwen2vl} & clean spans & training set \\
gaze, head-rhythm & $\tilde{\phi}$~\cite{pan2024s3} & per-frame pseudo-labels (pooled to tokens) & behavior control \\
\bottomrule
\end{tabular}}
\end{table}
\subonly{\vspace{-0.5em}}
\par \noindent\textbf{Behavior labels.}
We obtain gaze and head-rhythm labels in two steps. First, we apply the kinematic decomposition of Pan~et~al.~\cite{pan2024s3} to the tracked facial-motion state, which combines expression, head pose, and eye-gaze estimates from the trackers described above. This yields smoothed partner-direction cues and pitch/yaw/roll motion statistics. Second, we discretize these signals into gaze labels (contact / avert) and head-rhythm labels (still / nod / shake / tilt) using thresholding with hysteresis; ambiguous frames are masked out of the behavior supervision.
The per-token control $\behavior_{k}$ is a mask-weighted average of the $4$ frames within each latent token's span, re-normalized to a simplex per channel; a two-bit validity mask $\mathbf{M}_{k} \in \{0,1\}^{2}$ marks gaze and head-rhythm independently as present or absent for that token.
These labels define the control vocabulary the user steers at inference and supervise the behavior observers during training. They are reproducible kinematic proxies, not ground-truth social-intent labels. \rev{Exact thresholds, hysteresis rules, aggregation, and filtering are given in the supplemental.}
The final filtered corpus contains $10{,}555$ clips, split $9{,}790$ train and $765$ test. %

\subsection{Latent Motion Prior} \label{sec:motion_repr}
Given the behavioral labels from Sec.~\ref{sec:data}, we learn a partner-aware and steerable prior over target motion. To reduce the temporal sequence length, we model motion in a learned latent space: a causal motion VAE compresses each $T{=}100$-frame window into $K{=}25$ latent tokens, and a causal flow-matching transformer models the distribution of these tokens conditioned on audio, partner context, style, and semantic controls. The VAE is trained first and then kept frozen for all downstream prior training.
\par \noindent\textbf{Motion autoencoder.}
A causal temporal VAE~\cite{kingma2014vae} encodes $T{=}\Tframes$-frame windows ($4$\,s at $25$\,fps) into $K{=}\Klatent$ latent tokens of dimension $D{=}\Dlatent$. The encoder $\AEenc$ produces a Gaussian posterior $q_\psi(\motionlatent_{1:K}\mid\motion^{1:T})$, sampled via the standard reparameterization,
\setlength{\abovedisplayskip}{2pt}\setlength{\belowdisplayskip}{2pt}%
\begin{equation} \label{eq:ae}
(\boldsymbol{\mu}_{1:K},\,\boldsymbol{\sigma}_{1:K}) = \AEenc(\motion^{1:T}), \qquad \motionlatent_{1:K} = \boldsymbol{\mu}_{1:K} + \boldsymbol{\sigma}_{1:K}\odot\boldsymbol{\epsilon}, \quad \boldsymbol{\epsilon}\sim\mathcal{N}(\mathbf{0},\mathbf{I});
\end{equation}
the decoder $\AEdec$ reconstructs $\hat{\motion}^{1:T}=\AEdec(\motionlatent_{1:K})$. Both networks are causal. The VAE is trained under
\begin{equation} \label{eq:ae_loss}
\mathcal{L}_{\mathrm{AE}} = \lambda_{\mathrm{rec}}\bigl\|\motion - \hat{\motion}\bigr\|_{1} + \lambda_{\mathrm{vel}}\bigl\|\Delta\motion - \Delta\hat{\motion}\bigr\|_{1} + \lambda_{\mathrm{KL}}\,D_{\mathrm{KL}}\!\bigl(q_\psi(\motionlatent\mid\motion)\,\|\,\mathcal{N}(\mathbf{0},\mathbf{I})\bigr),
\end{equation}
where $\Delta$ denotes a one-frame finite difference. Once converged, the VAE is frozen and all downstream modeling operates on its latent space. We sample from the Gaussian posterior during training of the prior and use the posterior mean $\boldsymbol{\mu}_{1:K}$ at deployment. %
\par \noindent\textbf{Causal flow-matching prior.} \label{sec:flow_prior}
On top of the frozen motion autoencoder, we learn a causal latent-space prior following MotionStreamer~\cite{xiao2025motionstreamer}. Unlike MotionStreamer, we use rectified-flow matching~\cite{liu2022rectifiedflow,lipman2023flowmatching} instead of a diffusion sampler, and condition the target motion on the target audio, partner audio, partner motion, and STEER's semantic controls.
Concretely, we learn a conditional flow-matching prior $\prior$ over target latents $\targetlatent$ given a set of conditioning inputs
\[
\mathbf{c} = \{\targetaudio,\ \partnercontext,\ \stylevec,\ \behavior_{1:K},\ \emotion\},
\]
where $\targetaudio$ is the target's own audio feature stream, $\partnercontext = \partnerenc(\partnerlatent,\ \partneraudio)$ is the Partner Encoder output over the partner's motion latents $\partnerlatent = \AEenc(\motion_{\mathrm{partner}})$ and audio features $\partneraudio$ ($\motion_{\mathrm{partner}}$ being the partner-side counterpart of $\motion$ from Eq.~\ref{eq:motion_vec}), $\stylevec$ is a motion-style code, and $\behavior_{1:K}, \emotion$ are the two semantic controls (per-token behavior, window-level emotion; Sec.~\ref{sec:semantic_controls}).
Sampling a flow-time variable $\tau \sim \mathcal{U}(0,1)$ and forming the rectified-flow interpolate
\begin{equation} \label{eq:rectified}
\motionlatent_{\tau} = (1-\tau)\,\boldsymbol{\epsilon} + \tau\,\targetlatent,\qquad \boldsymbol{\epsilon}\sim\mathcal{N}(\mathbf{0},\mathbf{I}_{K \times D}),
\end{equation}
we regress the velocity field with the flow-matching objective
\begin{equation} \label{eq:flow}
\mathcal{L}_{\mathrm{flow}} = \mathbb{E}_{\tau,\boldsymbol{\epsilon},(\targetlatent,\mathbf{c})}\Bigl[\bigl\|\prior(\motionlatent_{\tau}, \tau, \mathbf{c}) - (\targetlatent - \boldsymbol{\epsilon})\bigr\|_{2}^{2}\Bigr].
\end{equation}
The network $\prior$ is a causal DiT~\cite{peebles2023dit} with RoPE self-attention~\cite{su2021roformer}.
\par \noindent\textbf{Partner Encoder.} $\partnerenc$ is a causal RoPE encoder that fuses $\partnerlatent$ and $\partneraudio$ into the per-token context $\partnercontext$.
\par \noindent\textbf{Conditioning pathways.}
The prior receives the target audio, partner context, motion style, behavior controls, and emotion label through modality-specific conditioning paths: cross-attention for audio and partner context, FiLM modulation \cite{perez2018film} for style, the behavior residual of Sec.~\ref{sec:behavior_ctrl}, and AdaLN-Zero for emotion. To maintain streaming continuity, we prepend a clean prefix of $n_{\mathrm{prev}}{=}3$ previously committed latents to the current candidate tokens. Audio features are extracted with a frozen causal Wav2Vec2 encoder~\cite{baevski2020wav2vec2,chen2025dystream}; conditioning dropout follows classifier-free guidance~\cite{ho2022cfg}. \subonly{Partner audio and partner motion are dropped jointly so that the partner stream is either fully present or fully replaced by its null embedding.}\rev{Partner audio and partner motion share a joint dropout mask that replaces the partner stream as a unit. Each partner modality is additionally dropped independently (supp.), so the model also trains on partner-audio-only and partner-motion-only states.}

\subsection{Semantic Control Injection} \label{sec:semantic_controls}
We inject each control through the mechanism best suited to its temporal rate and value type: FiLM modulation~\cite{perez2018film} for the clip-level continuous style code, a gap-driven residual on the last three DiT layers for the per-token discrete behavior controls, and AdaLN-Zero~\cite{peebles2023dit} for the window-level discrete emotion class.
\par \noindent\textbf{Style.} \label{sec:style_ctrl}
We additionally condition the prior on a compact target-reference code $\stylevec \in \mathbb{R}^{\Dstyle}$, encoded from a same-speaker motion window, to capture residual target-specific motion statistics~\cite{sun2024diffposetalk,danecek2023emote,fan2022faceformer,peng2023emotalk}. A KL-regularized VAE limits the capacity of this code so that dyadic dynamics must still be explained by the audio and partner streams. The code modulates the flow-matching backbone via per-block FiLM; when no reference is available, we use a learned null code.
\par \noindent\textbf{Behavior.} \label{sec:behavior_ctrl}
The user steers the target with per-token control vectors stacking a $2$-class gaze softmax (contact / avert) and a $4$-class head-rhythm softmax (still / nod / shake / tilt),
\begin{equation} \label{eq:behavior_vec}
\behavior_{k} = [\,\mathbf{q}_{k}^{g};\ \mathbf{q}_{k}^{h}\,] \in \mathbb{R}^{\Dbehav},\qquad \mathbf{q}_{k}^{g} \in \Delta_{2},\ \mathbf{q}_{k}^{h} \in \Delta_{4},
\end{equation}
where $[\,\cdot\,;\,\cdot\,]$ denotes channel-wise concatenation; the format permits independent specification across the two channels.\footnote{We write $\Delta_{n}$ for the $n$-vertex probability simplex, i.e.\ non-negative vectors in $\mathbb{R}^{n}$ summing to one.}
\par \noindent\textit{Gap-driven residual.}
Behavior labels often agree with what the audio-and-partner backbone already produces (e.g. a contact request when the target is already looking at the partner); concatenating $\behavior_{1:K}$ as input tokens would force the prior to re-predict that behavior even when no change is needed. We instead inject the \emph{gap} between requested and currently-predicted behavior as a residual on the last three DiT layers. At each such layer, an observer $\observer_{\ell}$ predicts $\hat{\behavior}_{\ell} = \observer_{\ell}(\mathbf{h}_{\ell})$ from the layer's hidden state, supervised by a masked cross-entropy loss $\mathcal{L}_{\mathrm{obs}}$ against the per-token labels $\behavior_{1:K}$ of Sec.~\ref{sec:data}. The control residual at layer $\ell$ is then
\begin{equation} \label{eq:behav_resid}
\mathbf{h}_{\ell}^{\mathrm{ctrl}} = \mathbf{h}_{\ell} + \mathcal{R}_{\ell}\!\bigl([\,\mathbf{h}_{\ell}\,;\ (\behavior_{1:K} - \hat{\behavior}_{\ell})\odot\mathbf{M}_{1:K}\,]\bigr),
\end{equation}
where $[\,\cdot\,;\,\cdot\,]$ denotes channel-wise concatenation, $\mathcal{R}_{\ell}$ is a small MLP zero-initialized at the residual output\rev{, in the spirit of zero-initialized control adapters~\cite{zhang2023controlnet,xie2024omnicontrol,dai2024motionlcm}}, and $\mathbf{M}_{1:K}$ is the per-token validity mask of Sec.~\ref{sec:data}, broadcast across each channel and exposed at inference for per-channel masking.
\par \noindent\textbf{Emotion.} \label{sec:emotion_ctrl}
A window-level $\Demo$-class label $\emotion \in \Delta_{\Demo}$ (a soft distribution over the seven classes at training, from the window-aggregation step of Sec.~\ref{sec:data}; user-specified at inference) is injected through AdaLN-Zero~\cite{peebles2023dit}: at each block $\ell$ an adapter $\mathcal{G}_{\ell}^{\mathrm{emo}}$ produces $(\boldsymbol{\gamma}_{\ell},\boldsymbol{\beta}_{\ell},\boldsymbol{\alpha}_{\ell}) = \mathcal{G}_{\ell}^{\mathrm{emo}}(\emotion)$ that gate the block's residual sub-branches:
\begin{equation} \label{eq:adaln}
\mathbf{h}_{\ell+1} = \mathbf{h}_{\ell} + \boldsymbol{\alpha}_{\ell} \odot \mathcal{F}_{\ell}\!\bigl((1+\boldsymbol{\gamma}_{\ell}) \odot \mathrm{LN}(\mathbf{h}_{\ell}) + \boldsymbol{\beta}_{\ell}\bigr),
\end{equation}
where $\mathcal{F}_{\ell}$ is the block's self-attention or MLP sub-network. Training-time supervision on $\emotion$ is via the observer-tracking loss of Sec.~\ref{sec:train_distill}, evaluated on the model's decoded motion against the requested class.
\subsection{Training and Streaming Inference} \label{sec:train_distill}
\par \noindent\textbf{Phase 1: backbone prior.}
To prevent the prior from collapsing onto the semantic labels before the audio-and-partner backbone has converged, Phase 1 trains the backbone alone under
\begin{equation} \label{eq:phase1}
\mathcal{L}_{\mathrm{phase1}} = \lambda_{\mathrm{flow}}\mathcal{L}_{\mathrm{flow}} + \lambda_{\mathrm{energy}}\mathcal{L}_{\mathrm{energy}},
\end{equation}
where $\mathcal{L}_{\mathrm{flow}}$ is Eq.~\ref{eq:flow} and $\mathcal{L}_{\mathrm{energy}}$ is an asymmetric per-region energy regularizer
$\mathcal{L}_{\mathrm{energy}} = \sum_{d} \mathrm{ReLU}\!\bigl(\alpha\, E_{d}^{\mathrm{gt}} - E_{d}^{\mathrm{pred}}\bigr)$~\cite{nair2010relu},
with the per-region energy $E_{d}(\motion) = \sqrt{\frac{1}{T-1}\sum_{t=2}^{T} \|\Delta\motion_{t,d}\|_{2}^{2}}$ (RMS first-difference velocity), floor ratio $\alpha{=}0.9$, and $d \in \{\mathrm{exp},\mathrm{jaw},\mathrm{neck},\mathrm{eye}\}$. The predicted energy is computed on the clean-motion estimate $\hat{\motion}$ obtained by a one-step flow rollout from $(\motionlatent_{\tau},\tau)$ to $\tau{=}1$. The ReLU activates only when the prediction under-shoots ground truth.
\par \noindent\textbf{Phase 2: semantic steerability.}
Phase 2 keeps the backbone trainable at a reduced learning rate and activates the semantic-control pathways under
\begin{equation} \label{eq:phase2}
\mathcal{L}_{\mathrm{STEER}} = \lambda_{\mathrm{flow}}\mathcal{L}_{\mathrm{flow}} + \lambda_{\mathrm{emo}}\mathcal{L}_{\mathrm{emo}} + \lambda_{\mathrm{obs}}\mathcal{L}_{\mathrm{obs}},
\end{equation}
where $\mathcal{L}_{\mathrm{emo}}$ is the cross-entropy between the requested emotion class and the window-level logits of a frozen \emph{affect observer}~\cite{danecek2023emote,peng2023emotalk} --- a TCN we pre-train on LibreFace labels (supp.) --- applied to the decoded motion, and $\mathcal{L}_{\mathrm{obs}}$ is masked cross-entropy on the behavior observers $\{\observer_{\ell}\}$ of Sec.~\ref{sec:behavior_ctrl}.
\rev{Both phases additionally apply the flow-matching loss on the decoded motion, weighted by $\lambda_{\mathrm{decoded}}$, and penalize per-region velocity and acceleration with weights $\lambda_{\mathrm{vel}}$ and $\lambda_{\mathrm{acc}}$. Phase~2 adds a preservation loss with weight $\lambda_{\mathrm{preserve}}$: the backbone's predictions under the null condition are matched to a frozen Phase-1 copy, keeping the audio-and-partner pathway stable while the control pathways train. Coefficients are listed in the supplemental.}
\par \noindent\textbf{Few-step distillation for live inference.} \label{sec:distill}
For live deployment we distil the $10$-step teacher $\prior^{\teach}$ into a $2$-step student $\prior^{\stud}$ under the same streaming protocol used at deployment: at each emit the model takes the $n_{\mathrm{prev}}$-token clean prefix (Sec.~\ref{sec:flow_prior}) and predicts $K$ candidate tokens of which a leading commit-stride $s{=}1$ is committed; the trailing $K{-}s$ tokens overlap with the next emit's candidate slots and are blended with the next emit's predictions through a raised-cosine taper of width $8$ tokens. We use $s{=}1$ throughout, including all reported evaluations: each token spans $4$ frames, so one commit decodes a $160$\,ms chunk at $25$\,fps.
The student is trained with a trajectory-matching loss in the spirit of progressive distillation~\cite{salimans2022progressive},
\begin{equation} \label{eq:distill}
\mathcal{L}_{\mathrm{distill}} = \bigl\|\targetlatent^{\stud} - \targetlatent^{\teach}\bigr\|_{2}^{2},
\end{equation}
combined with the same semantic supervision applied to the teacher. The full per-emit budget at $s{=}1$ on an RTX $5090$ and the resulting cue-to-render latency are reported in the supplementary.
\subsection{Driving a Photoreal Gaussian Head Avatar} \label{sec:translator}
\begin{figure}[t]
\centering
\includegraphics[width=\linewidth]{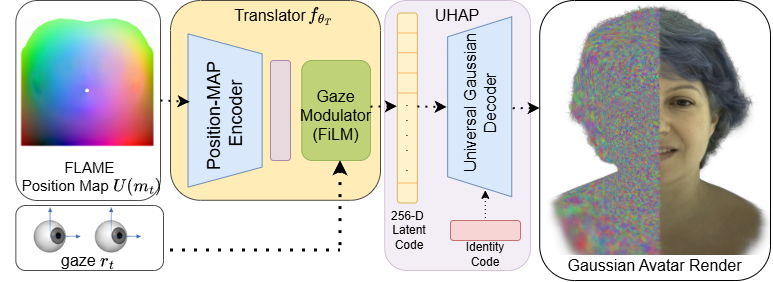}
\caption{Translator $f_{\theta}$. The FLAME geometry from $\motion_t$ --- with eye rotation neutralized --- is rendered as a UV position map $U(\motion_t)$ and encoded into UHAP's $\Davatar$-D expression latent; the eye-pose vector $\mathbf{r}_t$ enters \emph{only} via a FiLM modulator, so gaze is encoded exactly once. UHAP's universal Gaussian decoder is kept frozen.} %
\label{fig:translator_arch}
\end{figure}

At inference, the target latent $\hat{\targetlatent}$ sampled by $\prior$ (Sec.~\ref{sec:flow_prior}) is decoded by the frozen motion AE into per-frame FLAME state $\hat{\motion}^{1:T} = \AEdec(\hat{\targetlatent})$; the rest of this section drives a photoreal head avatar from those frames.
We drive UniGAHA's Universal Head-Avatar Prior (UHAP)~\cite{teotia2025unigaha}, whose Gaussian decoder takes a $\Davatar$-D expression code and an identity code. Gaze is entangled with the rest of UHAP's expression code, so a per-frame translator $f_{\theta}: \mathbb{R}^{\Dmotion} \to \mathbb{R}^{\Davatar}$ maps the FLAME~\cite{li2017flame} state $\motion_t$ into the expression code while taking gaze as an independent input (Fig.~\ref{fig:translator_arch}); the resulting disentangled control axes are visualised in Fig.~\ref{fig:flame_disentangle}.
\par \noindent\textbf{Architecture.}
The FLAME geometry --- with eye rotation held at the neutral pose, so gaze does not enter through the UV path --- is rendered as a UV position map $U(\motion_t)$ and fed to a position-map encoder $\mathcal{E}_{\mathrm{UV}}$. The eye-pose $\mathbf{r}_t$ enters \emph{only} through a gaze MLP $g_{\mathrm{gaze}}$ that emits FiLM parameters at the encoder bottleneck; a head $h_{\mathrm{lat}}$ projects the modulated features to the $\Davatar$-D expression code:
\begin{equation} \label{eq:translator}
f_{\theta}(\motion_t) = h_{\mathrm{lat}}\!\Bigl(\mathrm{FiLM}_{\boldsymbol{\gamma}_t,\boldsymbol{\beta}_t}\!\bigl(\mathcal{E}_{\mathrm{UV}}(U(\motion_t))\bigr)\Bigr), \qquad (\boldsymbol{\gamma}_t,\boldsymbol{\beta}_t) = g_{\mathrm{gaze}}(\mathbf{r}_t).
\end{equation}
UHAP's universal Gaussian decoder, identity embedding, and FLAME forward pass are kept frozen; only $f_{\theta}$ is trained under
\begin{equation} \label{eq:render_loss}
\mathcal{L}_{\mathrm{render}} = \lambda_{\mathrm{lat}}\mathcal{L}_{\mathrm{lat}} + \lambda_{\mathrm{vel}}\mathcal{L}_{\mathrm{vel}} + \lambda_{\mathrm{acc}}\mathcal{L}_{\mathrm{acc}} + \lambda_{\mathrm{pix}}\mathcal{L}_{\mathrm{pix}} + \lambda_{\mathrm{geo}}\mathcal{L}_{\mathrm{geo}},
\end{equation}
where $\mathcal{L}_{\mathrm{lat}}$ matches the translated code $f_{\theta}(\motion_t)$ to UHAP's pretrained expression latent, and $\mathcal{L}_{\mathrm{vel}}$ and $\mathcal{L}_{\mathrm{acc}}$ enforce first- and second-order temporal consistency in the same latent space. $\mathcal{L}_{\mathrm{pix}}$ distills UHAP's original encoder by matching renders driven by $f_{\theta}$ to renders driven by the pretrained UHAP encoder, while $\mathcal{L}_{\mathrm{geo}}$ preserves the positions and scales of the predicted Gaussians. At inference, the neck rotation $\mathbf{n}_t$ is applied as a rigid transform to the decoded Gaussians. \rev{Details are given in the supplemental.}

\section{Experiments} \label{sec:experiments}
We first describe the data, baselines, and metrics used throughout (Sec.~\ref{sec:datasets}), then present qualitative controllability results (Sec.~\ref{sec:qualitative_results}) and comparisons with state-of-the-art (Sec.~\ref{sec:comparisons}). Ablation studies isolate the partner-conditioning channels (Sec.~\ref{sec:ablations}); training-objective ablations are reported in the supplemental.
\subsection{Datasets, Baselines, and Metrics}
\label{sec:datasets}

\par \noindent\textbf{Data.}
We train STEER on the $9{,}790$-clip training split of our filtered RealTalk corpus (Sec.~\ref{sec:data}) and evaluate on a held-out test set of $765$ clips, partitioned by conversational role into a \textsc{Speak} subset ($n{=}170$, target purely speaking), a \textsc{Listen} subset ($n{=}330$, target purely listening), and a \textsc{Mixed} turn-taking subset ($n{=}265$, the target alternates between speaking and listening). The main-paper Tab.~\ref{tab:sota} reports the test-set average across all three subsets; the per-subset breakdown is in the supplemental.
All methods consume the partner stream (audio + motion) and the target's own audio. STEER runs in its base generation mode for this comparison --- the semantic controls are dropped to their classifier-free-guidance null tokens --- so every method generates target motion from the same partner+target audio inputs. Tab.~\ref{tab:sota} therefore reports the partner-input-to-motion mapping for each model under a matched input regime. The controllability eval --- whether STEER's semantic-control pathways produce the requested behavior when the controls \emph{are} supplied --- is reported separately in the supplemental.
\par \noindent\textbf{Baselines.}
We compare against two recent dyadic motion baselines fine-tuned on our corpus: DualTalk~\cite{peng2025dualtalk}, a joint speaker--listener model, and UniLS~\cite{chu2025unils}, an end-to-end audio-driven dyadic avatar.
We additionally include DiffPoseTalk~\cite{sun2024diffposetalk}, a monadic speech-driven baseline that does not condition on the partner, as a non-dyadic reference, and Ours-Distill, the $2$-step distilled student, alongside the full model. %
\par \noindent\textbf{Metrics.}
We adopt established metrics from the talking-head and listener-generation literature, organized along four axes; per-clip values are computed on the rendered FLAME mesh except where noted.
\emph{Lip fidelity.} \emph{LipSync\,(mm)} ($\downarrow$) reports the lip-vertex error in millimetres against ground truth, in the spirit of the lip-vertex error of~\cite{xing2023codetalker,sun2024diffposetalk}.
\emph{Realism and dynamics.} \emph{vFD\textsubscript{E}} ($\downarrow$) is the Fréchet distance between predicted and ground-truth motion sequences in expression-vertex space, following the L2L motion-realism protocol of Ng~\etal~\cite{ng2022learningtolisten}. \emph{FDD\textsubscript{u}}, \emph{PDD\textsubscript{h}}, and \emph{JDD} ($\downarrow$) are the \emph{Facial Dynamics Deviation}~\cite{xing2023codetalker} on the upper-face, head pose, and FLAME jaw axis-angle respectively --- the per-vertex gap between prediction and ground truth in temporal standard deviation --- which captures whether the predicted motion has the same level of dynamism as ground truth in each region.
\emph{Diversity.} \emph{Div\textsubscript{E}} and \emph{Div\textsubscript{H}} (bits, $\uparrow$) are the Shannon-index diversity of Ng~\etal~\cite{ng2022learningtolisten}: with k-means cluster centers pre-fit on ground-truth expression and head-rotation sequences, we report the entropy of the cluster-assignment histogram of the predicted samples, so that a uniform spread across modes scores higher than a collapsed prediction.
\emph{Partner-conditioned reactivity.}
\emph{RPCC\textsubscript{n}} ($\downarrow$) follows~\cite{song2023elp,peng2025dualtalk} and measures the $\ell_1$ gap between predicted and ground-truth target--partner Pearson correlations on the neck axis-angle.
\emph{rPCC\textsubscript{f\,v}} ($\downarrow$, $\times 10^{-4}$) applies the same residual-correlation metric to pose-normalized facial vertex motion induced by expression and jaw.

\subsection{Qualitative Results} \label{sec:qualitative_results}

Fig.~\ref{fig:controllability_roles} evaluates controllability across the two conversational roles. Under four identical control inputs (\emph{base}, \emph{happy}, \emph{avert}, \emph{avert\,+\,disgust}), the target exhibits the requested gaze and emotion edits in both regimes while preserving role-specific articulation, such as listening head motion and speaking lip motion.
Fig.~\ref{fig:flame_disentangle} further shows that these controls transfer to the photoreal avatar stage: gaze and expression edits remain visually disentangled after mapping the generated FLAME motion into UHAP's avatar-driving space.
\ifinlinefigs
\begin{figure*}[t]
\centering
\includegraphics[width=0.78\textwidth]{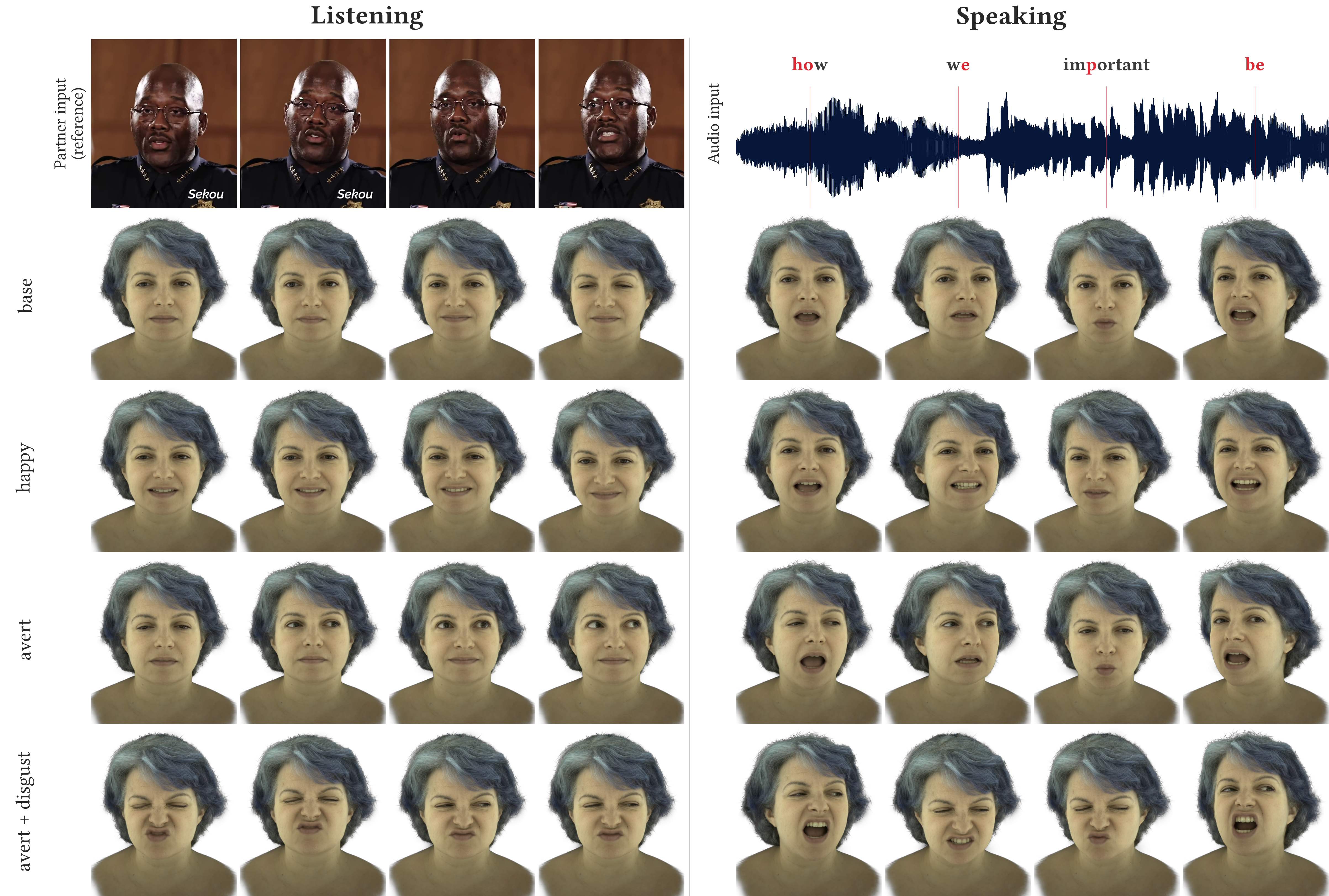}
\caption{Controllability across conversational roles. Top row: the dominant driving input per role (partner reference for \textsc{Listen}, target's own audio for \textsc{Speak}). Rows 2--5: same identity under four controls (\emph{base}, \emph{happy}, \emph{avert}, \emph{avert\,+\,disgust}); the controls take effect in both regimes while role-specific behavior (listening posture, speaking lip-sync) is preserved.}
\label{fig:controllability_roles}
\end{figure*}
\begin{figure}[t]
\centering
\includegraphics[width=\linewidth]{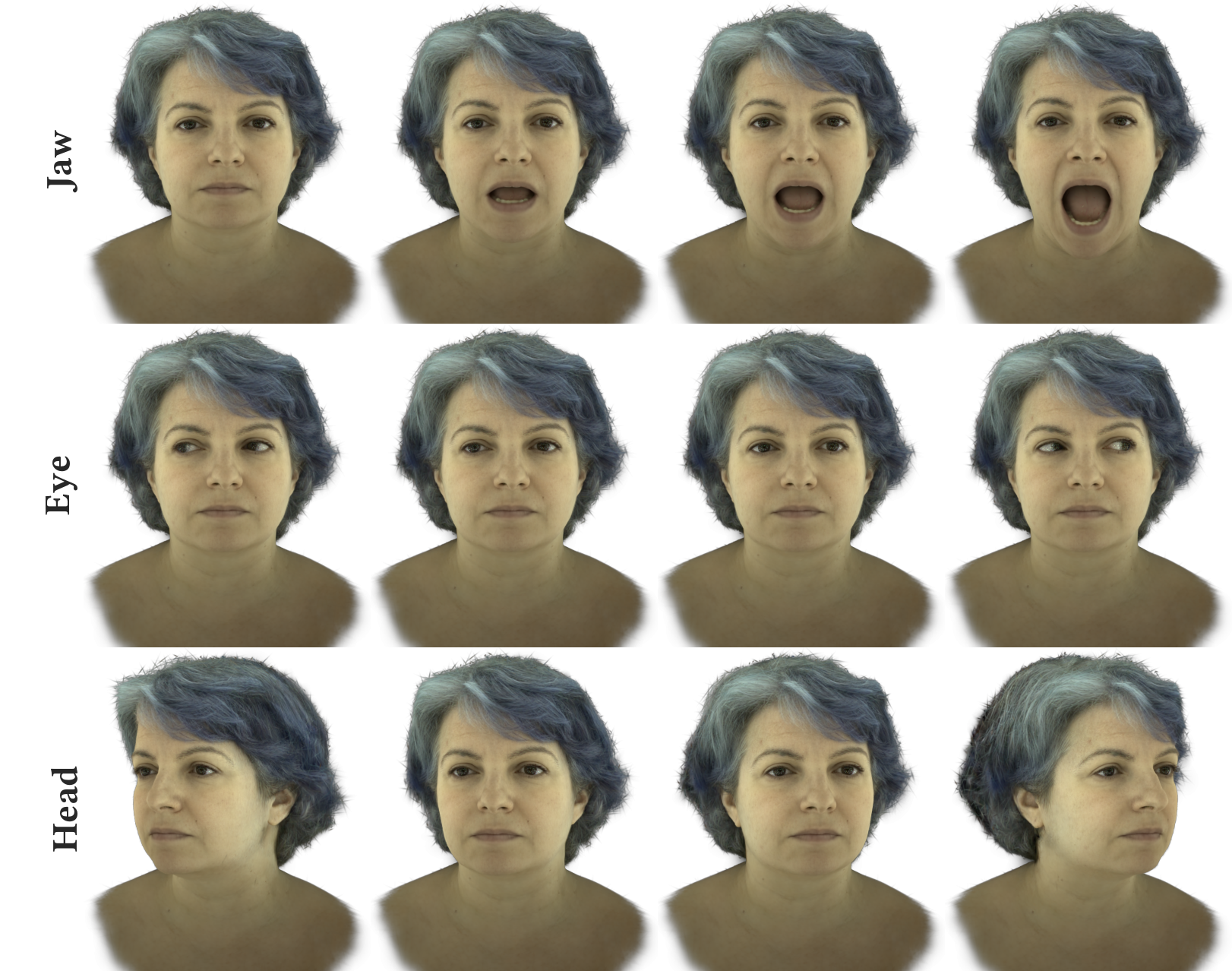}
\caption{Disentangled FLAME controls transferred to the Gaussian avatar through $f_{\theta}$. Independent sweeps of head yaw, eye yaw, and jaw opening produce clean, isolated changes in the rendered avatar while other motion channels remain fixed.}
\label{fig:flame_disentangle}
\end{figure}
\fi

\subsection{Comparisons with State-of-the-art Methods} \label{sec:comparisons}
\begin{table}[t]
\centering
\caption{Quantitative comparison on the held-out test set ($n{=}765$, weighted across the \textsc{Speak} / \textsc{Listen} / \textsc{Mixed} subsets defined in Sec.~\ref{sec:datasets}; per-subset breakdown in the supplemental). Each method generates target motion from the partner stream (audio + motion) and target audio; STEER runs in its base mode with semantic controls dropped to the classifier-free-guidance null tokens, matching baselines' input regime. The orthogonal controllability evaluation under \emph{supplied} controls is in the supplemental. LipSync\,(mm) is the per-frame mean lip-vertex error; rPCC\textsubscript{f\,v} is reported as ${\times}10^{-4}$. Best per column in bold; cells color-shaded by per-column rank.}\label{tab:sota}
\setlength{\tabcolsep}{2pt}
\scriptsize
\resizebox{\linewidth}{!}{%
\begin{tabular}{@{}lccccccccc@{}}
\toprule
{Method} & {LipSync$\downarrow$} & {vFD\textsubscript{E}$\downarrow$} & {FDD\textsubscript{u}$\downarrow$} & {JDD$\downarrow$} & {PDD\textsubscript{h}$\downarrow$} & {Div\textsubscript{E}$\uparrow$} & {Div\textsubscript{H}$\uparrow$} & {RPCC\textsubscript{n}$\downarrow$} & {rPCC\textsubscript{f\,v}$\downarrow$} \\
\midrule
DiffPoseTalk~\cite{sun2024diffposetalk} & 12.46 & 0.0376 & 0.0405 & 0.1415 & 0.3029 & 1.246 & 0.723 & 0.481 & 8.86 \\
DualTalk-FT~\cite{peng2025dualtalk}     & 9.33 & 0.0572 & 0.0648 & 0.2011 & 0.3299 & 1.010 & 0.619 & \cellcolor{rank1}\textbf{0.397} & 6.12 \\
UniLS-FT~\cite{chu2025unils}            & 11.67 & 0.0394 & 0.0444 & 0.1357 & 0.3169 & 1.143 & 0.923 & 0.458 & 7.38 \\
Ours-Distill                            & \cellcolor{rank2}9.18 & \cellcolor{rank2}0.0267 & \cellcolor{rank2}0.0267 & \cellcolor{rank2}0.1026 & \cellcolor{rank2}0.2430 & \cellcolor{rank2}1.562 & \cellcolor{rank2}0.980 & 0.419 & \cellcolor{rank1}\textbf{5.62} \\
\textbf{Ours (Full)}                    & \cellcolor{rank1}\textbf{9.15} & \cellcolor{rank1}\textbf{0.0262} & \cellcolor{rank1}\textbf{0.0260} & \cellcolor{rank1}\textbf{0.1008} & \cellcolor{rank1}\textbf{0.2343} & \cellcolor{rank1}\textbf{1.595} & \cellcolor{rank1}\textbf{1.086} & \cellcolor{rank2}0.406 & \cellcolor{rank2}5.65 \\
\bottomrule
\end{tabular}}
\end{table}
\par \noindent\textbf{Quantitative.}
Tab.~\ref{tab:sota} summarizes the comparison on the full $n{=}765$ test set, averaged across \textsc{Speak} / \textsc{Listen} / \textsc{Mixed}; per-subset results are provided in the supplemental.
STEER is the strongest overall model, achieving the best scores on lip accuracy, motion realism, dynamics, and diversity.
The only metric not led by STEER is neck-level partner coupling, where DualTalk is marginally better, but this comes with noticeably reduced diversity.
UniLS uses partner audio but does not condition on partner motion, and remains below STEER across the reported metrics, while the monadic DiffPoseTalk baseline is weakest on partner-reactivity metrics.
The distilled student remains close to the full model, preserving the main quantitative gains while supporting real-time inference.
\par \noindent\textbf{Qualitative.}
Fig.~\ref{fig:qualitative} provides a visual comparison on representative \textsc{Speak} and \textsc{Listen} clips. With global head pose normalized, the upper-face region in our outputs contains eye-pose and brow motion that is largely absent in the baseline rows on \textsc{Speak}; on \textsc{Listen}, the target's smile response and head motion in our outputs follow the partner cue (top RGB row), whereas the baselines remain comparatively static.
\ifinlinefigs
\begin{figure*}[t]
\centering
\includegraphics[width=0.65\textwidth]{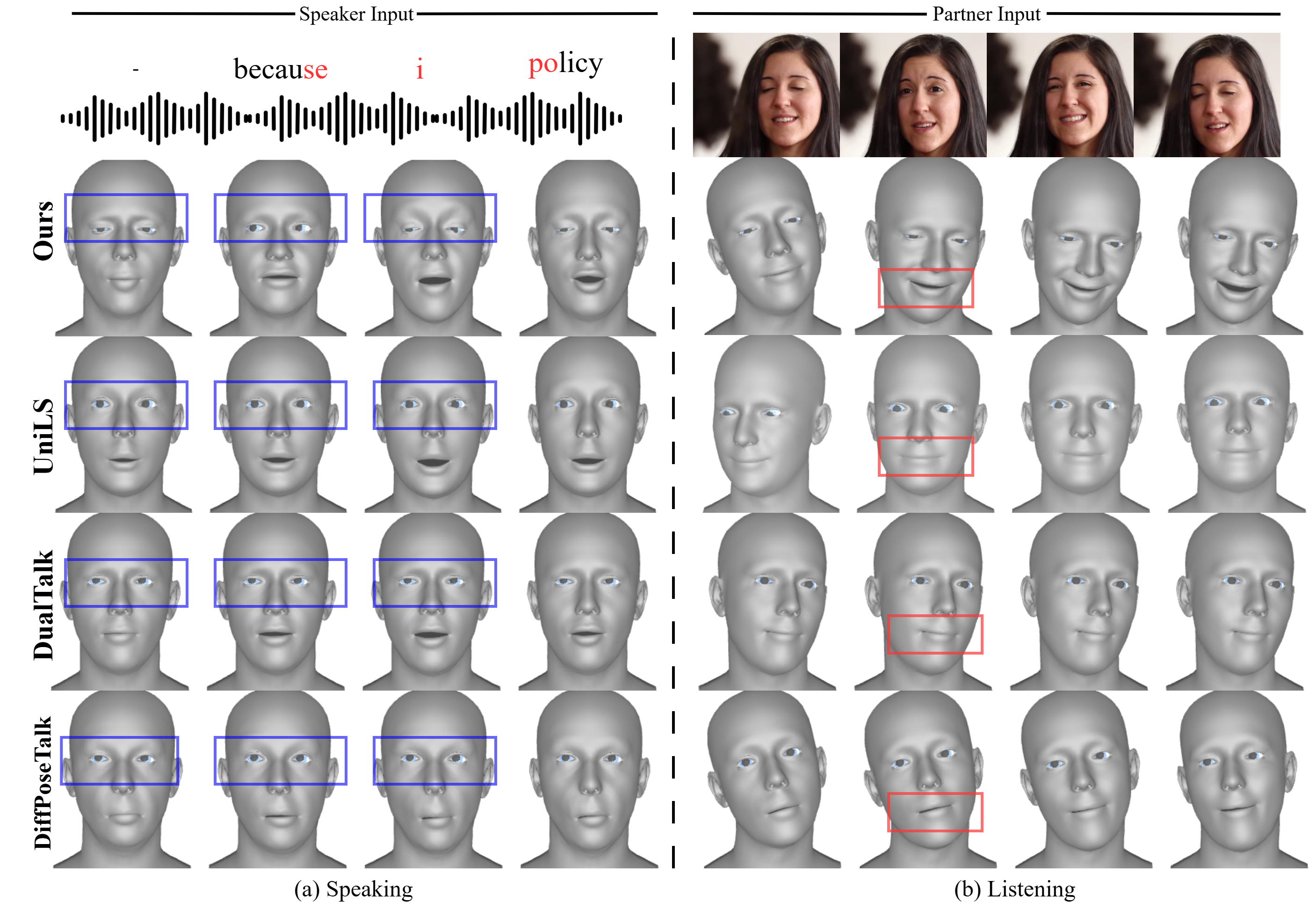}
\caption{Qualitative comparison; STEER (top) versus UniLS, DualTalk, DiffPoseTalk. \emph{(a) \textsc{Speak}}: blue boxes mark the upper face --- with head pose normalized, STEER shows richer eye-pose and brow motion while baselines stay static. \emph{(b) \textsc{Listen}}: red boxes mark the mouth; STEER mirrors the smiling partner (top, RGB) more strongly and varies head motion more than baselines.}
\label{fig:qualitative}
\end{figure*}
\fi
\rev{%
\subsection{Perceptual User Study} \label{sec:user_study}
In a blinded two-alternative forced-choice study, raters see STEER beside one baseline on $18$ held-out clips ($8$ \textsc{Listen}, $8$ \textsc{Speak}, $2$ \textsc{Mixed}), judging lip-sync (\textsc{Speak}), reaction timing (\textsc{Listen}/\textsc{Mixed}), expressions, and head pose (all clips); $10$ raters took part, and STEER is preferred on every facet against every baseline (Tab.~\ref{tab:user_study}).
\begin{table}[t]
\centering
\caption{\rev{Perceptual user study: share (\%) of pairwise comparisons in which raters preferred STEER over each baseline, per facet ($>\!50$ favors STEER). In each trial of the blinded two-alternative forced-choice study, raters saw STEER and one baseline side by side on the same clip ($10$ raters, $18$ held-out clips). Lip-sync is judged on \textsc{Speak} clips, reaction timing on \textsc{Listen} and \textsc{Mixed} clips, and expressions and head pose on all clips.}}
\label{tab:user_study}
\scriptsize
\setlength{\tabcolsep}{5pt}
\begin{tabular}{@{}lcccc@{}}
\toprule
{vs.\ STEER} & {Lip-sync} & {Expressions} & {Head pose} & {Reaction} \\
\midrule
UniLS        & $67$ & $71$ & $71$ & $85$ \\
DualTalk     & $72$ & $81$ & $84$ & $91$ \\
DiffPoseTalk & $85$ & $88$ & $90$ & $84$ \\
\midrule
All          & $76$ & $81$ & $83$ & $87$ \\
\bottomrule
\end{tabular}
\end{table}

}%

\subsection{Partner-conditioning Ablation} \label{sec:ablations}
The partner stream carries two channels --- partner audio and partner motion (latents) --- on top of the target's own audio. Fig.~\ref{fig:mirror_laugh} illustrates the redundancy this gives the model on a mirror-laugh clip: the smile response is elicited even when partner audio or partner motion alone is provided; both channels carry the social cue. Removing both partner channels jointly suppresses the response, confirming the partner stream as the source of the mirroring rather than the target's own audio. A detailed quantitative analysis of the partner-conditioning ablation is provided in the supplementary document.
\ifinlinefigs
\begin{figure}[t]
\centering
\includegraphics[width=\linewidth]{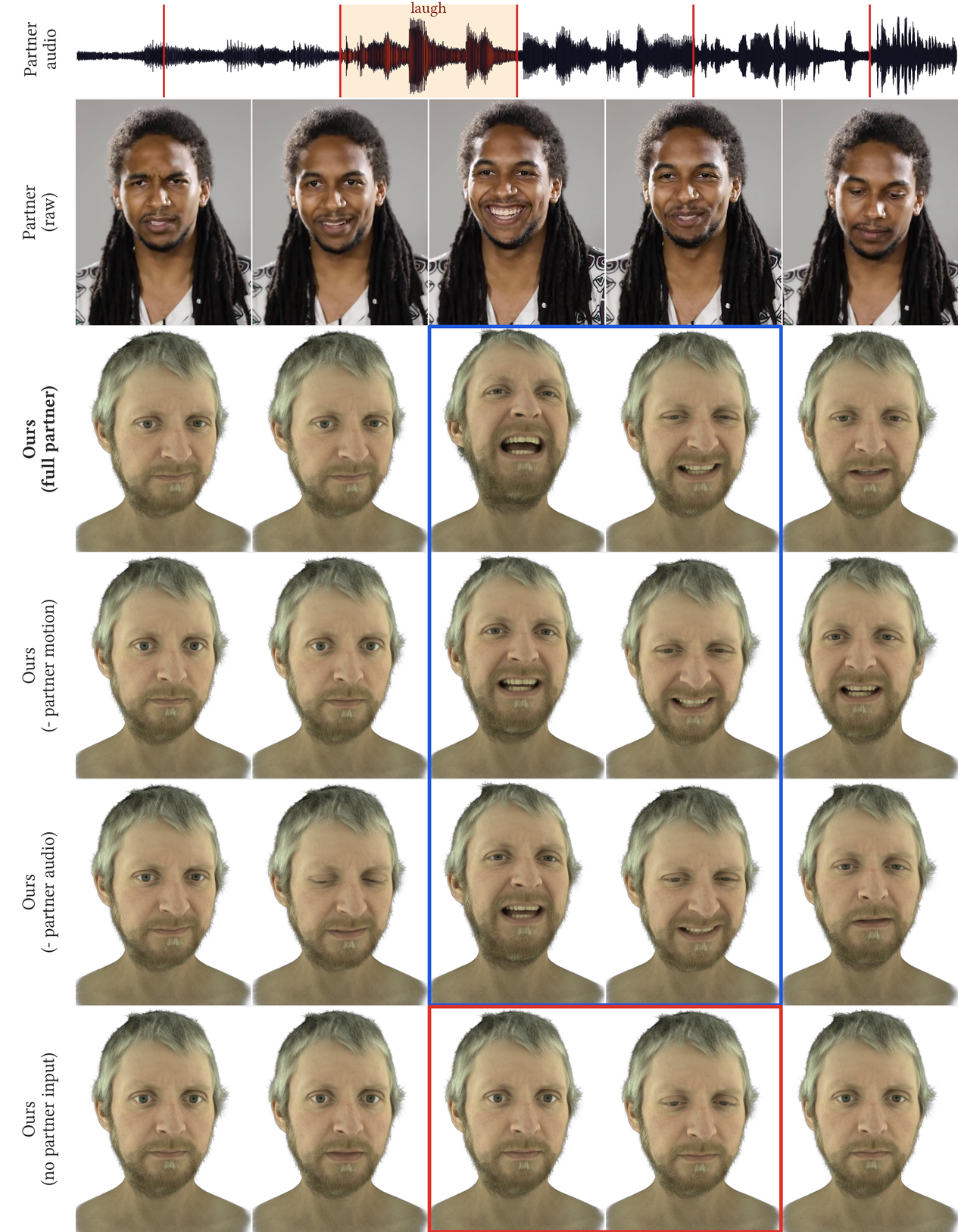}
\caption{Mirror-laugh ablation. Both partner channels carry the cue: partner audio or partner motion alone elicits the smile (rows 3--5, blue); removing both fails to elicit it (row 6, red).}
\label{fig:mirror_laugh}
\end{figure}
\fi
\subsection{Real-time System Deployment} \label{sec:live_system}
We additionally deploy STEER as an interactive avatar that runs end-to-end at $25$\,fps on a single NVIDIA RTX~$5090$~\cite{nvidia2025rtx5090}; the per-stage cue-to-render latency breakdown is reported in the supplemental.
The system can be interfaced in three input modes: \emph{text} (typed prompts driving the partner via Gemini 2.5 Flash~\cite{google2025gemini25flash} TTS), \emph{audio-only} (microphone input from the partner), and \emph{audio + video}.
At runtime the user can swap the rendered identity, change the camera viewpoint of the Gaussian avatar, and override the semantic controls; Fig.~\ref{fig:live_system} reports the running HUD across these axes.
We provide screen recordings of all three interaction modes, with on-screen control overlays and audio, in the supplemental video.
\ifinlinefigs
\begin{figure}[t]
\centering
\includegraphics[width=\linewidth]{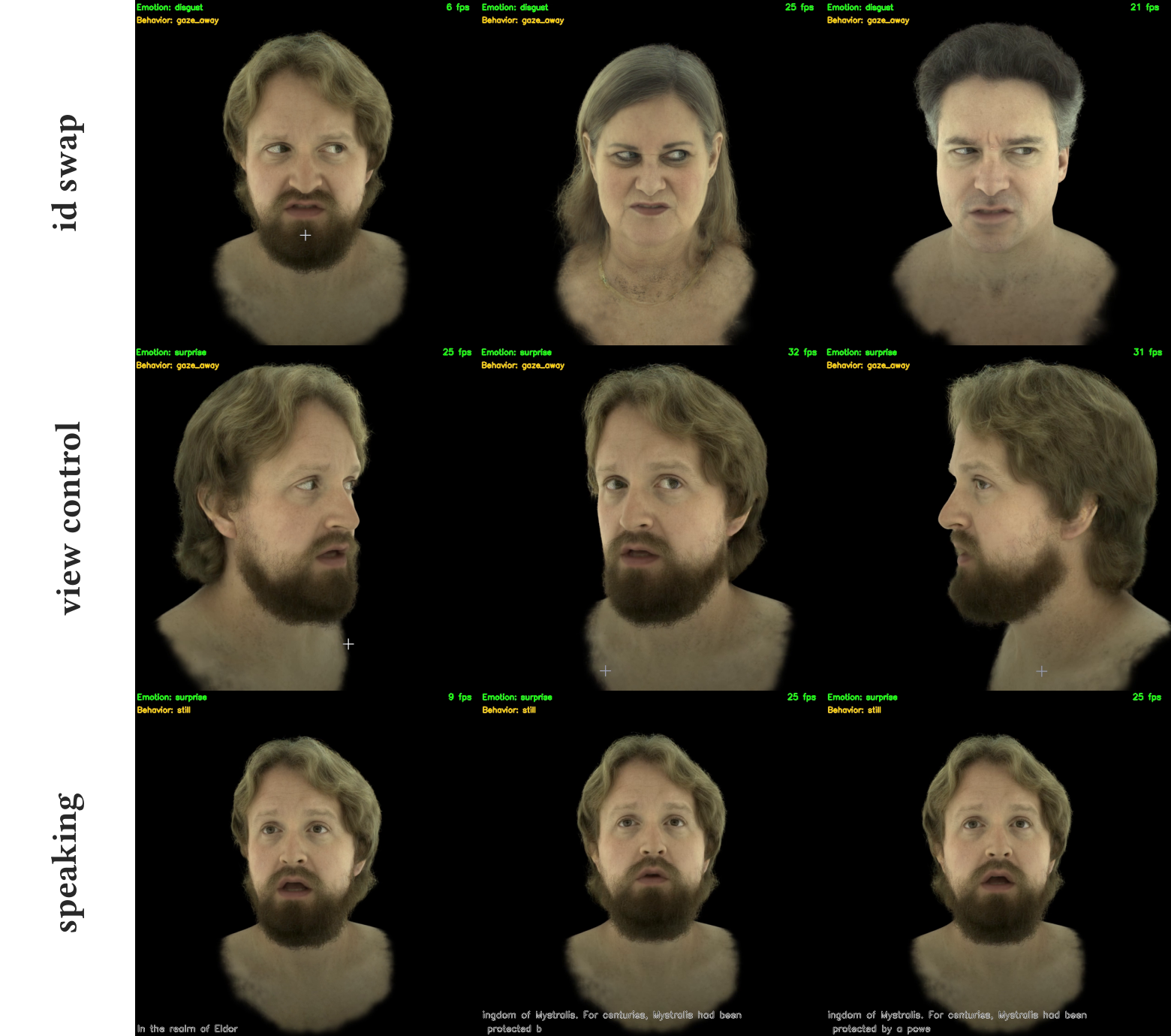}
\caption{STEER deployed in a real-time setting at $25$\,fps. \emph{Row 1 (id swap)}: same pose driving three different identities through UHAP. \emph{Row 2 (view control)}: free-viewpoint rendering of the same identity. \emph{Row 3 (speaking)}: live audio-driven speaking with semantic-control overlays (emotion, behavior).}
\label{fig:live_system}
\end{figure}
\fi

\section{Limitations and Future Work} \label{sec:discussion}
STEER exposes explicit semantic controls for gaze, head rhythm, and emotion, but these axes remain discrete by design: two gaze modes, four head-rhythm modes, and seven emotion classes. This does not capture the full continuous space of human expressive variation. \rev{The decomposition is deliberately interpretable, since controllability requires axes that a user can name and request; learned decompositions such as PCA or codebooks do not offer this.} Extending the controller to continuous affect coordinates, richer behavior vocabularies, or free-form text specifications is an important direction for future work.
Our interactive system runs in real time, but currently depends on a high-end GPU for the full avatar pipeline. \rev{In the live mode, motion tokens are committed before future context arrives, which can introduce high-frequency head motion that does not occur offline.} Improving model efficiency, renderer throughput, and deployment on more accessible hardware would make the system more practical for broader interactive applications.
STEER also assumes that the desired controls are provided externally. While this is useful for authoring and interactive steering, a natural next step is to add a higher-level behavior planner that selects gaze, head rhythm, and affect from conversational content, dialogue state, partner cues, or user intent.
Finally, the model inherits biases and failure modes from the underlying tracking and annotation stack, including the RealTalk data distribution and LibreFace-derived emotion supervision. \rev{Training on a single corpus also limits demographic coverage. Behavior also varies across subjects: the style code captures this only over short windows, and a subject-conditioned behavior handle is a natural extension.} Extending the data coverage, improving annotation robustness, coupling the controller to full-body motion, and learning personalized behavior preferences online are promising future directions.
\rev{\paragraph{Ethical considerations.}
Photoreal avatars can be misused for unauthorized depiction, privacy violations, and deceptive content; ours come from a studio-captured prior of consenting subjects, and we do not construct photoreal avatars of in-the-wild individuals: RealTalk video supervises motion, not appearance. Demographic bias from the tracking and labeling stacks remains a concern, and deployments should disclose synthesis and obtain consent.}

\section{Conclusion} \label{sec:conclusion}
We presented STEER, a controllable dyadic motion prior for photorealistic conversational head avatars. By exposing gaze, head rhythm, and emotion as explicit semantic controls, STEER moves beyond prior dyadic methods that leave non-verbal behavior largely implicit in audio and partner conditioning.
To enable this control, we introduced an annotation pipeline for in-the-wild dyadic video, trained a causal flow-matching prior for partner-aware target motion, and connected the generated motion to a universal Gaussian head-avatar prior through a learned FLAME-to-avatar driving-space translator. Experiments show that STEER improves motion quality, dynamics, and diversity over recent dyadic baselines, while supporting controllable edits and real-time interactive deployment. \rev{We will release our training and evaluation code together with our dataset annotations.}

\section{Acknowledgements}
We would like to thank Flawless AI for financially supporting Kartik Teotia for this work.

\bibliographystyle{ACM-Reference-Format}
\bibliography{references}

\clearpage
\raggedbottom   %
\begin{center}
{\LARGE\bfseries Supplementary Material}\\[0.6em]
{\large STEER: Steerable Dyadic Head Avatars}
\end{center}
\vspace{1em}
\appendix
\section{Behavior-Label Pipeline and Thresholds}
\label{sec:label_thresholds}

The gaze and head-rhythm pseudo-labels referenced in main-paper Sec.~3.1 are derived from the kinematic motion decomposition $\tilde{\phi}$ of Pan~et~al.~\cite{pan2024s3}, which produces per-frame partner-attention and rhythmic-head signals from the tracked facial-motion state, including head-pose and eye-gaze estimates. We turn those signals into discrete categorical labels in three stages: per-frame thresholding, per-token pooling, and clip-level filtering.

\paragraph{Geometry assumption.}
RealTalk shots are filmed face-to-face but the camera distance and framing vary, so we do not assume any fixed seating geometry. Instead we estimate a per-shot \emph{partner direction} for each speaker as a robust trimmed mean of their head-forward vector across the shot --- i.e., where the speaker is mostly looking on average. All gaze angles below are measured in degrees relative to this shot-specific reference.

\paragraph{Per-frame gaze (contact / avert).}
For each frame we measure the angle $\theta$ between the speaker's combined head + eye gaze vector and the partner direction. With hysteresis,
\begin{itemize}\setlength{\itemsep}{0pt}
\item $\theta < 11^\circ$ enters \textbf{on-partner} state (within a narrow cone toward the partner),
\item $\theta > 19^\circ$ leaves it; in between the state is held to avoid flicker.
\end{itemize}
Each frame is then either on-partner ($1$) or averted ($0$). To require temporal consistency, we smooth this $\{0,1\}$ signal over a $7$-frame ($\sim\!280$\,ms) window and label a frame \textbf{contact} if at least $5$ of those $7$ frames were on-partner, \textbf{avert} if at most $2$ were, and mask out the gaze channels in between (the ambiguous band).

\paragraph{Per-frame head-rhythm (still / nod / shake / tilt).}
We classify each frame from the rhythmic head-pose components (pitch, yaw, roll, all in degrees) over a local window centered on that frame, with half-window $hw{=}12$ frames ($\sim\!1$\,s total span). Let $\Delta_p, \Delta_y, \Delta_r$ be the within-window peak-to-peak range of pitch / yaw / roll, and $\bar{|\dot p|}$ the mean per-frame absolute pitch derivative. The per-frame label is:
\begin{itemize}
\item \textbf{still}: $\bar{|\dot p|} < 0.5^\circ/\text{frame}$ \emph{and} $\Delta_p < 1^\circ$ \emph{and} $\Delta_y < 1^\circ$;
\item \textbf{nod}: $\Delta_p > 1.2 \cdot \max(\Delta_y, \Delta_r)$;
\item \textbf{shake}: $\Delta_y > 1.2 \cdot \max(\Delta_p, \Delta_r)$;
\item \textbf{tilt}: $\Delta_r > 1.2 \cdot \max(\Delta_p, \Delta_y)$;
\item otherwise the head-rhythm channels are masked out at that frame..
\end{itemize}
The $1.2{\times}$ dominance multiplier suppresses noisy near-tied frames; the $0.5^\circ$ / $1^\circ$ stillness floors were calibrated by histogram inspection on $\sim\!200$ training shots.

\paragraph{Per-token pooling.}
Each $T{=}100$-frame window (4\,s at $25$\,fps) is reduced to $K{=}25$ latent tokens by the AE; we pool the per-frame labels onto the same grid by splitting the window into $25$ contiguous bins of $4$ frames. For each bin we take a mask-weighted average of the per-frame labels and re-normalize the gaze channels and the head-rhythm channels to a simplex; a channel group's bin is marked valid iff at least one frame in the bin was unmasked for that group. This yields the per-token control vector $\mathbf{b}_k$ and validity mask $\mathbf{M}_k$ used in main-paper Sec.~3.3.

\paragraph{Clip-level filtering.}
Before pseudo-labels are computed, we filter shots and windows for tracking quality: (i) a Qwen2-VL pre-pass~\cite{wang2024qwen2vl} flags shots with side-views, multi-person frames, or heavy occlusion (filtering at the shot level); (ii) windows in which the kinematic decomposition $\tilde{\phi}$ is missing or incomplete are dropped; (iii) only contiguous clean spans of at least $5$ seconds are retained for training. These steps reduce the $16{,}573$ raw RealTalk shots to the $10{,}555$-clip filtered corpus reported in the main paper. \rev{The pipeline itself is corpus-agnostic: it consumes tracked motion and per-speaker audio, so any face-to-face corpus with separable audio can be processed the same way.}

\section{Network Architectures}
\label{sec:network_arch}

This section gives the per-module specifications referenced from main-paper Sec.~3.2--3.5. All downstream modules operate at the token rate $K{=}\Klatent$ defined there unless noted.

\subsection{Motion autoencoder ($\AEenc, \AEdec$, main-paper Sec.~3.2)} \label{ssec:motion_ae}
The motion AE is a standard VAE~\cite{kingma2014vae}; its design closely follows the causal motion-latent autoencoder used by MotionStreamer~\cite{xiao2025motionstreamer} but is retrained on our $\Dmotion$-D dyadic FLAME state. The encoder $\AEenc$ stacks two stride-$2$ causal $1$D convolutions (kernel $5$, channels $\Dmotion{\to}128{\to}256$), followed by a $2$-layer Transformer with rotary position embeddings (RoPE)~\cite{su2021roformer}, $4$ heads, and feed-forward width $512$, which produces the per-token Gaussian-posterior parameters $(\boldsymbol{\mu}_k,\boldsymbol{\sigma}_k)$. The decoder $\AEdec$ mirrors this stack with transposed causal convolutions. The reconstruction loss in main-paper Eq.~3 is a per-region weighted $\ell_1$ ($w_{\mathrm{exp}}{=}1$, $w_{\mathrm{jaw}}{=}2$, $w_{\mathrm{neck}}{=}4$, $w_{\mathrm{eye}}{=}3$, $w_{\mathrm{eyelid}}{=}2$) plus first-order velocity $\ell_1$ ($\lambda_{\mathrm{vel}}{=}1.0$) and second-order acceleration $\ell_1$ ($\lambda_{\mathrm{acc}}{=}0.1$); the KL term toward $\mathcal{N}(\mathbf{0},\mathbf{I})$ is annealed from $0$ to $\beta_{\mathrm{KL}}{=}10^{-5}$ over $10{,}000$ iterations. The AE is trained for $150{,}000$ iterations at batch size $128$, lr $2{\times}10^{-4}$, and then frozen for all downstream modelling. \rev{Freezing the AE gives the prior a fixed latent distribution to model rather than a moving target, following the standard two-stage recipe in latent generative modeling~\cite{rombach2022ldm,chen2023mld,xiao2025motionstreamer}. A possible concern is that the bottleneck smooths away brief or subtle motion. In practice it does not: the smile-timing F1 (\S\ref{sec:partner_ablation_supp}) and the control effect-sizes of \S\ref{sec:control_compliance} are measured on motion decoded through this AE, and both depend on such events surviving decoding.}

\subsection{Partner Encoder ($\partnerenc$, main-paper Sec.~3.2)} \label{ssec:partner_enc}
The Partner Encoder turns the per-frame partner streams into the per-token context sequence $\partnercontext$ that every DiT block cross-attends to. Two streams enter: (i) AE-encoded partner motion latents $\partnerlatent$, already at the token rate; and (ii) partner audio features $\partneraudio$, mean-pooled over each $4$-frame token span from the audio encoder of \S\ref{ssec:audio}. Each stream is linearly projected to the shared hidden dimension and concatenated, followed by a $\textsc{Linear}{+}\textsc{GELU}{+}\textsc{Linear}$ fusion. The result is processed by a Transformer with RoPE~\cite{su2021roformer}, causal in time, $4$ blocks deep, hidden dimension $448$, $8$ heads, MLP ratio $4$. The output sequence is the partner KV consumed by the DiT.

\subsection{Flow-matching prior ($\prior$, main-paper Sec.~3.2--3.3)} \label{ssec:dit}
The prior is a causal DiT~\cite{peebles2023dit} with RoPE self-attention~\cite{su2021roformer}: $8$ blocks, hidden $448$, $8$ heads (head dimension $56$), MLP ratio $4$. Each block sequentially applies, in order: AdaLN-Zero modulation on the residual stream from the window-level emotion code $\emotion$~\cite{peebles2023dit}, causal RoPE self-attention over the $K{+}n_{\mathrm{prev}}$ tokens, cross-attention to the target's audio features $\targetaudio$, cross-attention to the Partner-Encoder output $\partnercontext$ (\S\ref{ssec:partner_enc}), per-block FiLM~\cite{perez2018film} modulation from the style code $\stylevec$, and a causal MLP. The AdaLN-Zero adapter $\mathcal{G}_\ell^{\mathrm{emo}}$ is a $2$-layer MLP from the $\Demo{=}7$-class emotion vector to per-block scale/shift/gate $(\boldsymbol{\gamma}_\ell,\boldsymbol{\beta}_\ell,\boldsymbol{\alpha}_\ell)$, with $\boldsymbol{\alpha}_\ell$ initialised to zero so the adapter starts as a no-op~\cite{peebles2023dit}. The style FiLM adapter is a single linear projection from $\stylevec \in \mathbb{R}^{128}$ to per-block $(\boldsymbol{\gamma},\boldsymbol{\beta})$.
\par \noindent\textbf{Gap-driven residual ($\mathcal{R}_\ell$).} The behavior residual of main-paper Sec.~3.3 (main-paper Eq.~7) lives on the last three DiT blocks. Each $\mathcal{R}_\ell$ is a $2$-layer MLP, hidden $448$, GELU, with its final linear layer zero-initialised so the residual is a no-op at training start. Each of the three layers carries its own observer head $\observer_\ell$ --- a single linear projection from the hidden state to per-channel gaze and head-rhythm softmaxes --- supervised by the masked cross-entropy of main-paper Sec.~3.4.

\subsection{Style VAE (main-paper Sec.~3.3)} \label{ssec:style_vae}
A VAE~\cite{kingma2014vae} encodes a single $T$-frame target-motion window into a clip-level motion-style code $\stylevec \in \mathbb{R}^{128}$. The encoder is a $1$D-conv stem followed by a $2$-layer Transformer, hidden $256$, $4$ heads, that pools to a single token and emits $\boldsymbol{\mu},\boldsymbol{\sigma} \in \mathbb{R}^{128}$. We sample at training and use $\boldsymbol{\mu}$ at deployment. The KL term toward $\mathcal{N}(\mathbf{0},\mathbf{I})$ caps the channel's capacity, as discussed in the main paper.

\subsection{Affect observer (main-paper Sec.~3.4)} \label{ssec:affect_observer}
The frozen affect observer that produces $\mathcal{L}_{\mathrm{emo}}$ is a causal $5$-block dilated TCN over per-frame motion $\motion \in \mathbb{R}^{T\times\Dmotion}$, hidden width $192$, kernel $5$, GroupNorm + GELU. Two heads share the trunk: a per-frame linear head outputs $12$-D action-unit intensities; a window head temporal-mean-pools the trunk features over $T$ frames and projects to $7$-class facial-expression logits. We pre-train it for $40{,}000$ steps with batch size $128$ on (motion, LibreFace) pairs from our training corpus, with a weighted sum of per-frame MSE on action units and window-level cross-entropy on the seven-class facial-expression label. The observer is frozen for all of STEER's training, and only the window-level emotion head is consumed by $\mathcal{L}_{\mathrm{emo}}$; the action-unit head is retained as auxiliary supervision during the observer's own pre-training.

\subsection{Audio encoders} \label{ssec:audio}
The target's and partner's audio are encoded by a shared, frozen Wav2Vec2-Large~\cite{baevski2020wav2vec2} adapted to the causal-streaming recipe of DyStream~\cite{chen2025dystream}: the lower convolutional feature extractor is unchanged; the top $6$ Transformer layers are made causal with a $60$\,ms right context. A $1024$-D audio adapter projects the Wav2Vec2 output to the DiT cross-attention dimension. The per-frame embeddings are mean-pooled over the $4$-frame token span before they enter the prior; the same encoder serves both the target-audio and partner-audio cross-attention paths in the DiT.

\subsection{Avatar translator $f_\theta$ (main-paper Sec.~3.5)} \label{ssec:translator}
\begin{sloppypar}
$f_\theta$ maps the per-frame FLAME state $\motion_t \in \mathbb{R}^{\Dmotion}$ to UHAP's $\Davatar$-D expression code. The FLAME geometry is first posed with expression, jaw, and neck applied and \emph{eye rotation held at the neutral pose}, so gaze does not enter the UV path; it is then rasterised into a $1024{\times}1024{\times}3$ UV position map $U(\motion_t)$ indexed by UV. This map feeds an $8$-layer weight-normalised 2D-conv encoder $\mathcal{E}_{\mathrm{UV}}$ with kernel $4$, stride $2$, LeakyReLU($0.2$); the channel progression is $32{\to}32{\to}64{\to}64{\to}128{\to}128{\to}256{\to}256$, downsampling $1024{\to}4$ spatially. The $256{\times}4{\times}4{=}4096$-D bottleneck is flattened and projected by two weight-normalised linear layers ($4096{\to}512{\to}\Davatar$) with LeakyReLU between, the last layer initialised at $0.1{\times}$. Gaze enters as an independent control through a $2$-layer MLP $g_{\mathrm{gaze}}$ ($4{\to}64{\to}2{\cdot}\Davatar$) that emits FiLM~\cite{perez2018film} parameters $(\boldsymbol{\gamma}_t,\boldsymbol{\beta}_t) \in \mathbb{R}^{\Davatar}$. The final code is
\[
f_\theta(\motion_t) \;=\; h_{\mathrm{lat}}\bigl(\mathcal{E}_{\mathrm{UV}}(U(\motion_t))\bigr) \,\odot\, (1{+}\boldsymbol{\gamma}_t) \,+\, \boldsymbol{\beta}_t.
\]
The gaze MLP's output layer is zero-initialised, so $f_\theta$ starts as a pure UV-only encoder. UHAP's Gaussian decoder, identity embedding, and FLAME forward pass are kept frozen.
\end{sloppypar}

\rev{%
\paragraph{Reconstruction fidelity.}
We measure what the translator costs relative to the avatar's own reconstruction ceiling on a held-out expression set ($3{,}000$ frames) with multi-view ground truth. The ceiling drives UHAP's frozen decoder with the expression codes recovered by its native encoder; the translator path drives the same decoder with $f_\theta$ applied to the tracked FLAME state. Tab.~\ref{tab:translator_fidelity_supp} reports both against the captured images. The translator lands within $3.8$\,dB PSNR of the ceiling. The residual gap originates primarily in the underlying FLAME tracking, which holds back on some subtle and extreme expressions before the avatar is driven. The gap is dominated by a small geometric misalignment ($+2.7$\,mm mesh error) rather than by a change in appearance (LPIPS $+0.044$). Fig.~\ref{fig:tracking_limitation} illustrates this: where the tracked FLAME state already lacks an expression, both the avatar's own reconstruction and the translator render inherit the loss.

\begin{figure}[H]
\centering
\includegraphics[width=\linewidth]{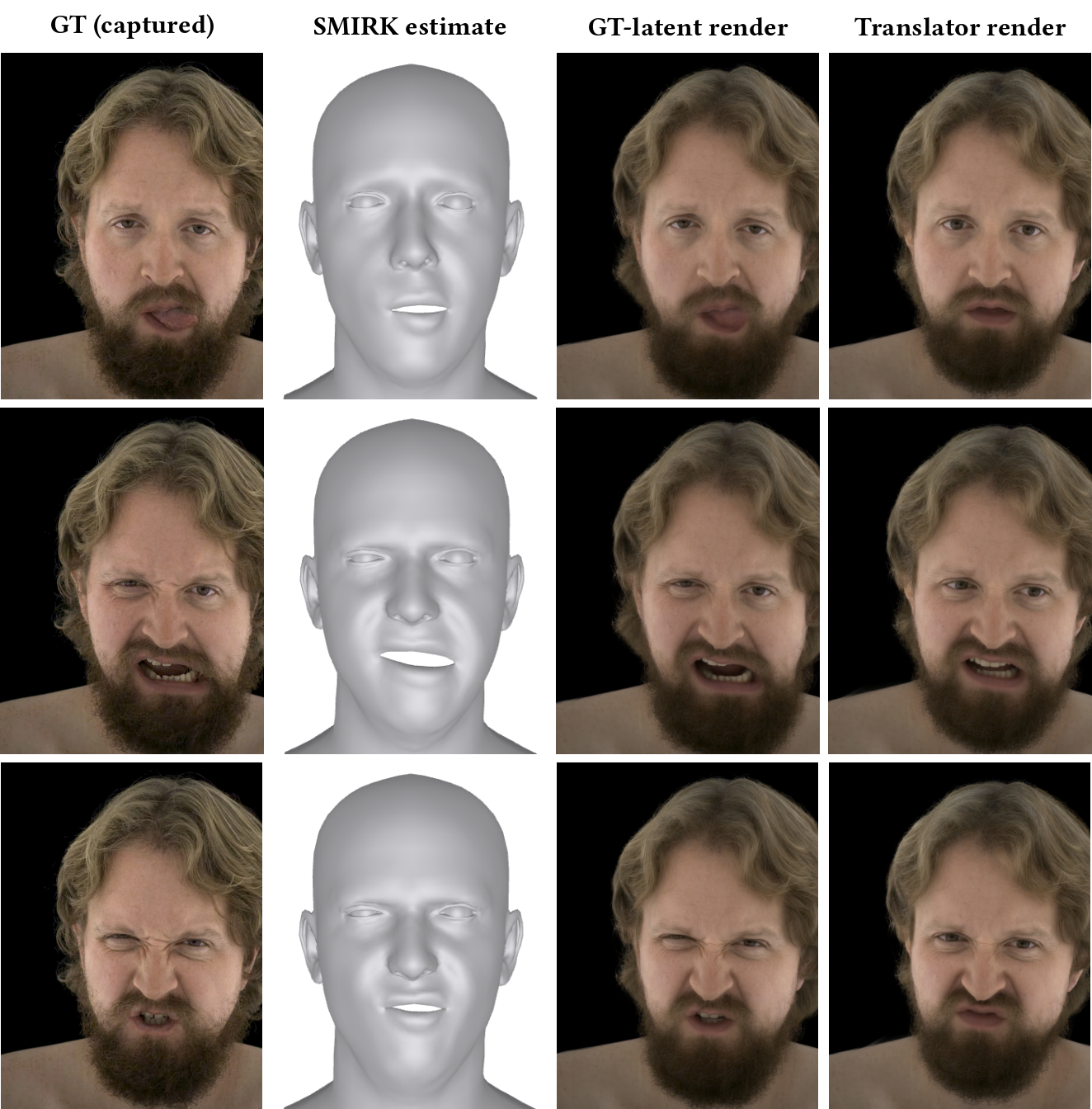}
\caption{\rev{Attribution of the avatar-bridge gap. Left to right: captured frame; tracked SMIRK/FLAME estimate (the translator's input); render from UHAP's native expression codes (the avatar's reconstruction ceiling); render from the translator. The tracked state already misses parts of the expression: the tongue in row 1, and fine expression detail in rows 2 and 3, most visibly the nose wrinkle in row 3. Both renders inherit this loss, so the residual gap originates in tracking, before the translator.}}
\label{fig:tracking_limitation}
\end{figure}

\begin{table}[H]
\footnotesize
\centering
\caption{\rev{Translator fidelity on a held-out expression set.}}
\label{tab:translator_fidelity_supp}
\setlength{\tabcolsep}{5pt}
\begin{tabular}{@{}lcccc@{}}
\toprule
 & {PSNR\,$\uparrow$} & {$\ell_1$\,$\downarrow$} & {LPIPS\,$\downarrow$} & {Mesh $\ell_2$\,$\downarrow$} \\
\midrule
GT-latent  & $28.05$ & $0.023$ & $0.165$ & $1.09$\,mm \\
Translator & $24.27$ & $0.032$ & $0.209$ & $3.79$\,mm \\
\midrule
Gap        & $-3.78$ & $+0.010$ & $+0.044$ & $+2.70$\,mm \\
\bottomrule
\end{tabular}
\end{table}
}%

\subsection{Preservation distillation} \label{ssec:preserve}
Fine-tuning $\prior$ jointly with the style, behavior, and emotion heads on top of an already-converged audio-and-partner backbone risks trunk drift. We mitigate this by holding a frozen copy of the backbone-only checkpoint as a \emph{preservation teacher} and adding a per-iteration trajectory-matching loss
\[
\mathcal{L}_{\mathrm{preserve}} = \bigl\|\targetlatent^{\stud, \emptyset} - \targetlatent^{\teach, \emptyset}\bigr\|_2^2,
\]
where both forward passes use the same noise/timestep but with \emph{all} semantic controls dropped out (the null condition). We weight $\mathcal{L}_{\mathrm{preserve}}$ at $1.0$ and lower the trunk learning rate to $6{\times}10^{-6}$, an order of magnitude below the per-head rates (\S\ref{ssec:training_schedule}). This keeps the audio-and-partner pathway aligned with its pre-fine-tune behaviour while the controls activate without overriding it.

\subsection{Training schedule} \label{ssec:training_schedule}
We optimise the prior with AdamW. Param-group learning rates: trunk $6{\times}10^{-6}$, audio adapter $1{\times}10^{-4}$, behavior heads $1.2{\times}10^{-4}$, style and emotion adapters $1.5{\times}10^{-4}$. We use cosine decay to $0.02{\times}$ of base LR with linear warmup, gradient-norm clipping at $1.0$, batch size $48$, and $215{,}000$ iterations. 

\par \noindent\textbf{Classifier-free guidance dropouts.} Following classifier-free guidance~\cite{ho2022cfg}, each conditioning channel is replaced with a learned null token at training with the per-channel Bernoulli probabilities of Tab.~\ref{tab:cfg_dropouts_supp}. The joint partner mask shares a single coin across partner audio and partner motion so the partner stream is replaced as a unit; the per-modality partner dropouts apply on top of that joint mask.

\begin{table}[H]
\footnotesize
\centering
\caption{Per-channel classifier-free-guidance dropout probabilities.}
\label{tab:cfg_dropouts_supp}
\setlength{\tabcolsep}{4pt}
\begin{tabular}{@{}ll@{}}
\toprule
Channel & Bernoulli $p$ \\
\midrule
Target audio                & $0.10$ \\
Partner stream (joint)      & $0.10$ \\
Partner audio / motion (per-modality)        & $0.05$ each \\
Style                       & $0.15$ \\
Behavior                    & $0.25$ \\
Emotion                     & $0.30$ \\
Previous-window prefix      & $0.10$ \\
\bottomrule
\end{tabular}
\end{table}

\par \noindent\textbf{Loss weights.}
Tab.~\ref{tab:loss_weights_supp} lists the loss coefficients used to instantiate the composite training loss of main-paper Sec.~3.4. The per-region $\ell_1$ velocity and acceleration losses use the FLAME-state regions $\{$neck, exp, jaw, eye$\}$ (acceleration on neck / exp / jaw only). The energy-floor losses use floor ratio $\alpha{=}0.90$ under a ReLU~\cite{nair2010relu} hinge.

\begin{table}[H]
\footnotesize
\centering
\caption{Loss coefficients (composite training loss, main-paper Sec.~3.4).}
\label{tab:loss_weights_supp}
\setlength{\tabcolsep}{4pt}
\begin{tabular}{@{}lll@{}}
\toprule
{Term} & {Symbol} & {Weight} \\
\midrule
Flow matching                    & $\lambda_{\mathrm{flow}}$    & $1.0$ \\
Decoded-motion flow              & $\lambda_{\mathrm{decoded}}$ & $1.0$ \\
Velocity $\ell_1$ (neck/exp/jaw/eye)   & $\lambda_{\mathrm{vel}}$     & $3.0 / 0.8 / 1.5 / 1.2$ \\
Accel.\ $\ell_1$ (neck/exp/jaw)        & $\lambda_{\mathrm{acc}}$     & $0.10 / 0.03 / 0.03$ \\
Energy floor (neck/exp/jaw/eye)        & $\lambda_{\mathrm{energy}}$  & $0.60 / 0.35 / 0.10 / 0.15$ \\
Preservation distillation        & $\lambda_{\mathrm{preserve}}$& $1.0$ \\
Behavior observer (hidden/decoded) & $\lambda_{\mathrm{obs}}$   & $0.30 / 0.08$ \\
Emotion observer (decoded)       & $\lambda_{\mathrm{emo}}$     & $0.50$ \\
\bottomrule
\end{tabular}
\end{table}

\subsection{Few-step distillation and streaming inference} \label{ssec:distill}
The $10$-step teacher is distilled to a $2$-step student in the spirit of progressive distillation~\cite{salimans2022progressive} (main-paper Eq.~11), trained with the same semantic supervision as the teacher so the control surface is preserved. At deployment the student runs streaming under the protocol of main-paper Sec.~3.4: at each emit it consumes the $n_{\mathrm{prev}}{=}3$-token clean prefix, predicts $K{=}25$ candidate tokens, commits $s{=}1$ leading token, and blends the trailing $K{-}s$ tokens into the next emit through a raised-cosine taper of width $8$ tokens. Keys and values from the self-attention and from both cross-attention paths are cached across emits, so the per-step compute scales with the commit stride rather than with the absolute time index.

\section{Per-Axis Behavior Controllability}

\begin{figure*}[!t]
\centering
\includegraphics[width=0.85\textwidth]{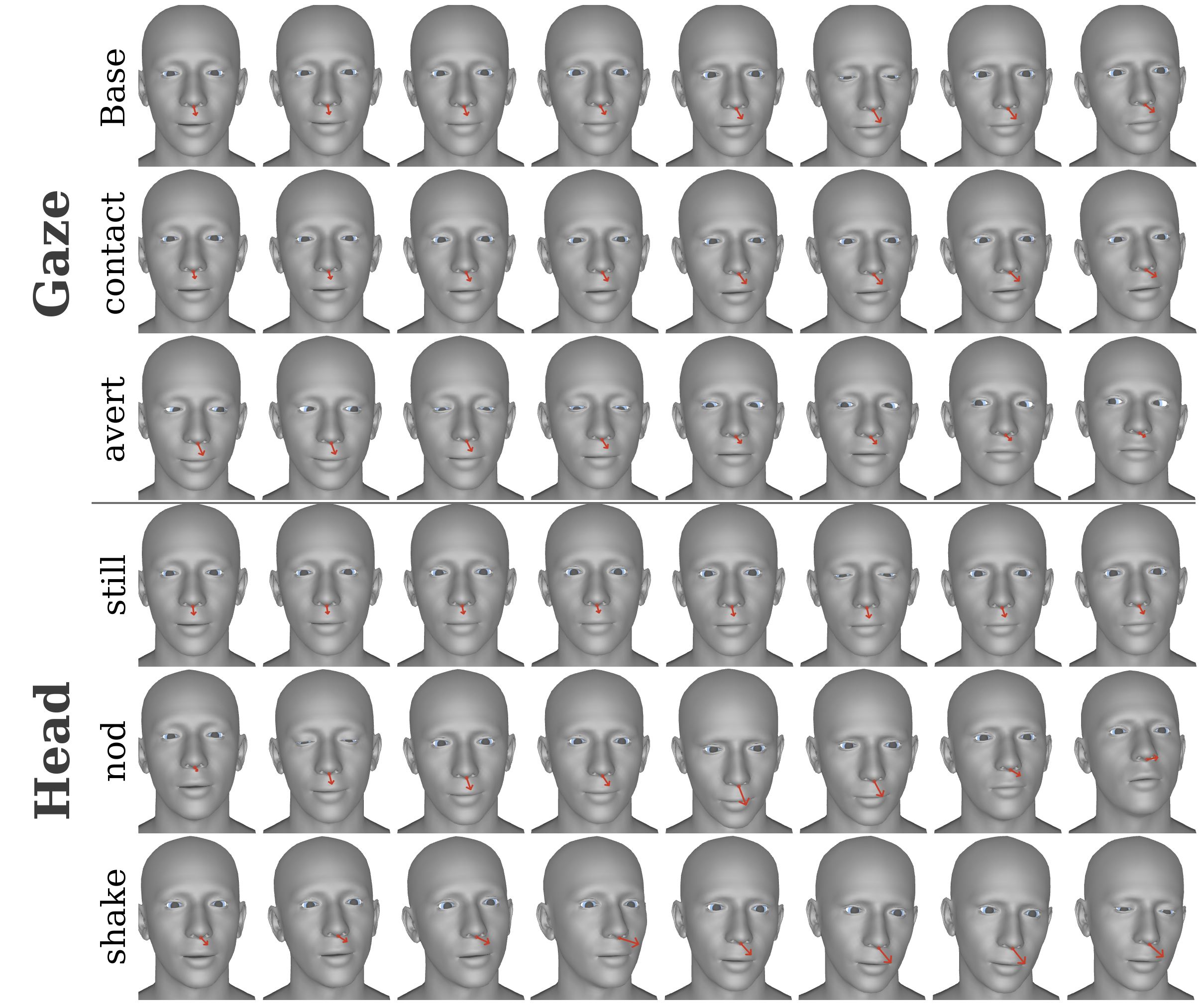}
\caption{Behavior-control showcase on a single test clip. \emph{Gaze} rows (Base, contact, avert) and \emph{Head} rows (still, nod, shake) are stacked vertically; columns are eight frames across time.}
\label{fig:edit_behavior_supp}
\end{figure*}

Fig.~\ref{fig:edit_behavior_supp} shows the per-axis behavior edits on a single test clip. The \emph{Gaze} axis (Base, contact, avert) selects between maintained gaze and gaze aversion; the \emph{Head} axis (still, nod, shake) selects between three head-rhythm modes. Changes are localized to the requested axis: head-mode edits do not alter gaze, and gaze edits do not alter head rhythm.

\section{Emotion Controllability}

\begin{figure*}[!t]
\centering
\includegraphics[width=0.85\textwidth]{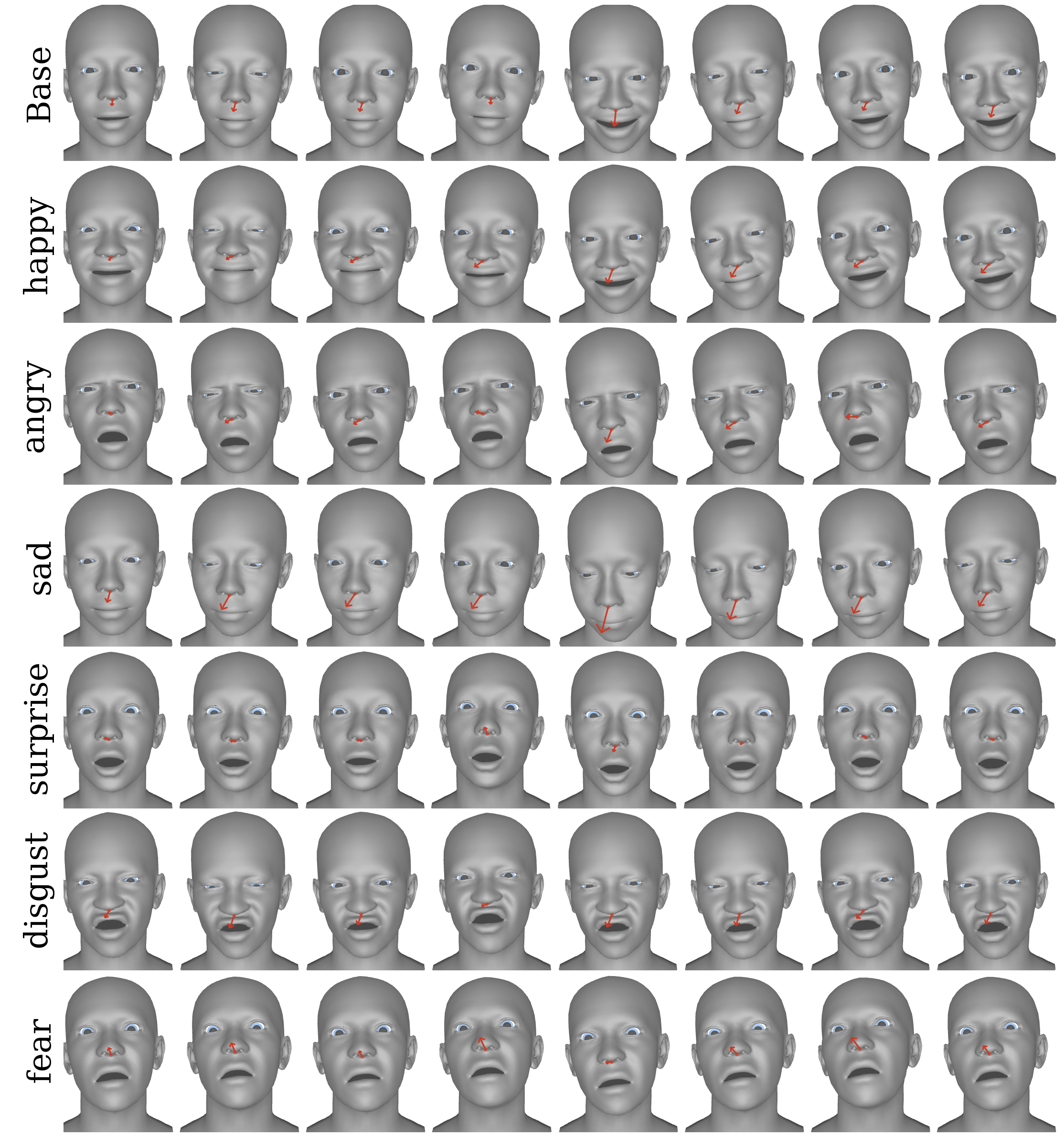}
\caption{Emotion-control showcase on a single test clip. Rows: \emph{Base}, \emph{happy}, \emph{angry}, \emph{sad}, \emph{surprise}, \emph{disgust}, \emph{fear}; columns are eight frames across time.}
\label{fig:edit_emotion_supp}
\end{figure*}

Fig.~\ref{fig:edit_emotion_supp} shows the same clip under the seven emotion classes. Each class consistently colors the target's expression while preserving the underlying motion structure shared with the base row, demonstrating that the emotion axis is orthogonal to the behavior controls of Fig.~\ref{fig:edit_behavior_supp}.

\section{Quantitative Control Compliance}
\label{sec:control_compliance}

The main-paper Tab.~2 evaluates motion quality with the semantic controls fixed to their ground-truth per-frame values, so the conditioning pathways themselves are not stressed by that table. To measure how reliably STEER obeys controls when they \emph{vary}, we re-use the same observers that supervised those pathways during Phase~2 training (main-paper Sec.~3.4) as test-time judges: each observer turns the generated motion back into a per-frame label distribution, and we compare the distribution under a requested control against the one the model produces on the same audio with all semantic controls dropped (the \emph{unconditioned} baseline). This tests exactly the contract the model was trained to satisfy, and avoids the methodological circularity of training a fresh classifier from the same supervision signal.

\paragraph{Protocol.}
We sample $n{=}100$ random clips from the \textsc{Listen} subset of the held-out test split, with the target's own audio silenced so the model is in pure-listener mode. For each clip we generate $14$ versions: an \emph{unconditioned} baseline; the $2$ gaze classes (\emph{contact}, \emph{avert}); the $4$ head-rhythm classes (\emph{still}, \emph{nod}, \emph{shake}, \emph{tilt}); and the $7$ emotion classes. Inference uses $10$-step rectified-flow sampling with audio classifier-free-guidance scale $1.5$, behavior-conditioning scales of $1.5$ (gaze) and $2.5$ (head-rhythm), and an emotion-strength multiplier of $1.2$.

\paragraph{Gaze.}
The gaze observer is the geometric contact/avert decision of Sec.~\ref{sec:label_thresholds}, applied per-frame to the predicted head-plus-eye vector against the partner direction with the same hysteresis used at training. Under conditioning, per-frame argmax matches the requested class on $95.4\%$ of frames for \emph{contact} and $91.5\%$ for \emph{avert} (Tab.~\ref{tab:cc_gaze_supp}). The model also produces a $1.19{\times}$ larger eye-pose magnitude under \emph{avert} than under \emph{contact}, so the conditioning produces the physical motion the observer is keying on rather than only its preferred label. Cross-axis leakage is small: gaze conditioning shifts the head-rhythm and emotion observer distributions by an average $L_1$ of below $0.03$ across both axes.

\begin{table}[H]
\footnotesize
\centering
\caption{Gaze compliance over $n{=}100$ \textsc{Listen} clips. Accuracy is per-frame argmax of the gaze observer (Sec.~\ref{sec:label_thresholds}) against the requested class. Off-axis $\Delta$ is the average $L_1$ shift the gaze conditioning induces in the head-rhythm / emotion observer distributions.}
\label{tab:cc_gaze_supp}
\setlength{\tabcolsep}{6pt}
\begin{tabular}{@{}lcccc@{}}
\toprule
{Requested} & {Acc.\ ($\uparrow$)} & {$\Delta_{\mathrm{rhythm}}$ ($\downarrow$)} & {$\Delta_{\mathrm{emo}}$ ($\downarrow$)} & {Eye-mag ratio} \\
\midrule
contact & $0.954$ & $0.032$ & $0.020$ & \multirow{2}{*}{$1.19{\times}$} \\
avert   & $0.915$ & $0.025$ & $0.028$ & \\
\bottomrule
\end{tabular}
\end{table}

\paragraph{Head-rhythm.}
For the three dynamic head-rhythm classes (\emph{nod / shake / tilt}) we report two effect-size statistics that measure whether the model amplifies motion on the requested neck axis (Tab.~\ref{tab:cc_rhythm_supp}). \emph{Range ratio} is the conditioned-generation range on the requested neck axis divided by the unconditioned-generation range on the same axis; \emph{axis dominance} is the requested-axis range divided by the maximum of the other two neck-axis ranges. Across $100$ clips, the requested axis is amplified by $1.91$--$2.20{\times}$ over the unconditioned baseline and becomes the dominant neck axis with margins of $1.22$--$1.66{\times}$.

\begin{table}[H]
\footnotesize
\centering
\caption{Head-rhythm compliance over $n{=}100$ \textsc{Listen} clips. Range ratio and axis dominance are effect-size statistics on the requested neck axis, defined inline.}
\label{tab:cc_rhythm_supp}
\setlength{\tabcolsep}{6pt}
\begin{tabular}{@{}lcc@{}}
\toprule
{Requested} & {Range ratio ($\uparrow$)} & {Axis dom.\ ($\uparrow$)} \\
\midrule
nod   & $1.91{\times}$ & $1.66{\times}$ \\
shake & $2.15{\times}$ & $1.22{\times}$ \\
tilt  & $2.20{\times}$ & $1.57{\times}$ \\
\bottomrule
\end{tabular}
\end{table}

\paragraph{Emotion.}
The emotion observer is our affect observer of \S\ref{ssec:affect_observer}, applied window-wise to the generated motion. Under emotion conditioning, the requested class receives the highest observer probability across all seven categories (Tab.~\ref{tab:cc_emotion_supp}). The target-emotion probability rises from its unconditioned prior on quiet listening clips --- $0.65$ for neutral, $0.35$ for happy, and below $0.001$ for the remaining five --- to $\geq\!0.999$ under the matching request. The peak action-unit activation also increases from $0.41$ unconditioned to $0.50$--$0.76$ under emotion conditioning, with the largest peaks under \emph{surprise} and \emph{fear} ($0.58$ and $0.76$), consistent with their canonical AU patterns. This suggests that the conditioning changes the generated facial motion in the intended direction, rather than only shifting an abstract classifier score.

\begin{table}[H]
\footnotesize
\centering
\caption{Emotion compliance over $n{=}100$ \textsc{Listen} clips. Accuracy is window-level argmax of our affect observer (\S\ref{ssec:affect_observer}) against the requested class; $p^{\mathrm{cond}}$ and $p^{\mathrm{uncond}}$ are the target-class probabilities under conditioning and on the unconditioned baseline; AU-peak$^{\mathrm{cond}}$ is the maximum action-unit activation on the conditioned generation.}
\label{tab:cc_emotion_supp}
\setlength{\tabcolsep}{6pt}
\begin{tabular}{@{}lcccc@{}}
\toprule
{Requested} & {Acc.\ ($\uparrow$)} & {$p^{\mathrm{cond}}$ ($\uparrow$)} & {$p^{\mathrm{uncond}}$} & {AU-peak$^{\mathrm{cond}}$} \\
\midrule
neutral   & $1.000$ & $\geq\!0.999$ & $0.650$  & $0.32$ \\
happy     & $1.000$ & $\geq\!0.999$ & $0.350$  & $0.50$ \\
sad       & $1.000$ & $\geq\!0.999$ & $<\!0.001$ & $0.51$ \\
surprise  & $1.000$ & $\geq\!0.999$ & $<\!0.001$ & $0.58$ \\
angry     & $1.000$ & $\geq\!0.999$ & $<\!0.001$ & $0.54$ \\
disgust   & $1.000$ & $\geq\!0.999$ & $<\!0.001$ & $0.53$ \\
fear      & $1.000$ & $\geq\!0.999$ & $<\!0.001$ & $0.76$ \\
\bottomrule
\end{tabular}
\end{table}

\rev{%
\paragraph{Independent estimators.}
The observers above share their supervision source with training. We therefore also verify the gaze and emotion controls with estimators that share no training signal with our pipeline: Py-Feat~\cite{cheong2023pyfeat} (emotion; AffectNet-trained) in place of our LibreFace-derived supervision, and L2CS-Net~\cite{abdelrahman2023l2cs} (gaze; Gaze360-trained) in place of our MediaPipe-derived eye pose. Both are applied to rendered video of the generated motion over $40$ held-out \textsc{Listen} clips under the same $14$-condition protocol (behavior-conditioning scale $2.0$, emotion strength $1.2$); L2CS-Net is run on the photoreal avatar renders. Under gaze conditioning, \emph{avert} produces $1.82{\times}$ the eye-yaw deviation of \emph{contact} ($95\%$ CI $\pm0.24$), with avert exceeding contact on $91\%$ of clips. Under emotion conditioning, the requested class's action-unit signature rises relative to the neutral condition on $95$--$100\%$ of clips for every class; the categorical readout recovers \emph{happy} and \emph{disgust} on $100\%$ of clips, \emph{neutral} on $95\%$, \emph{angry} on $92\%$, \emph{surprise} on $78\%$, \emph{sad} on $55\%$, and \emph{fear} on $28\%$; sad and fear are also the lowest-recall classes for AffectNet-trained classifiers on real faces. These independently-sourced measurements are consistent with the observer-based compliance above.
}%

\paragraph{Motion-quality preservation under conditioning.}
A control surface that satisfies the observer by destroying the rest of the face would not be useful in practice. Across all $14$ conditions we therefore monitor the velocity standard deviation of the jaw, upper-face, and head/neck vertex sets. Jaw velocity-std stays in $[0.001,0.002]$ across every condition --- so no spurious mouth motion is induced by any non-speaking control, despite the controls operating at full strength. The upper-face velocity-std rises by $\sim\!25\%$ under \emph{surprise} and \emph{fear} (expected: these emotions activate brow and lid regions), and the head/neck velocity-std rises by $\sim\!50\%$ under \emph{nod / shake / tilt} (also expected: those classes are by construction motion on the requested neck axis). The rest of the face is preserved under all controls.

\section{Inference Speed and Live Deployability}

\begin{table}[h]
\small
\centering
\caption{Inference speed and causality on a single NVIDIA L40 (model-forward only; audio I/O and rendering excluded). The \textsc{Causal} column indicates whether the model can be driven by streaming audio. Output motion is at $25$\,Hz.}
\label{tab:fps_supp}
\setlength{\tabcolsep}{8pt}
\begin{tabular}{@{}lrrc@{}}
\toprule
{Method} & {Params} & {FPS} & {Causal} \\
\midrule
DualTalk    & $647$M       & $6946$ & $\times$ \\
UniLS       & $352$M       & $1095$ & $\times$ \\
Ours ($10$-step)         & $\sim\!150$M & $46$   & \checkmark \\
Ours-Distill ($2$-step)  & $\sim\!150$M & $280$  & \checkmark \\
\bottomrule
\end{tabular}
\end{table}

Tab.~\ref{tab:fps_supp} reports motion frames per second on a single NVIDIA L40, measuring forward-pass wall time only (audio I/O, FLAME decoding, and rendering excluded); we use L40 here to match the hardware on which the baselines were profiled. DualTalk and UniLS report high offline throughput because their forward pass is non-causal --- each consumes the entire utterance in a single batched call --- which produces a high FPS number on this axis but is a different operating mode from streaming emission. STEER is strictly causal (a $60$\,ms audio-encoder look-ahead) and at the streaming protocol of main-paper Sec.~3.4 with commit stride $s{=}1$, each forward emit commits a single $4$-frame motion token ($160$\,ms of playback at $25$\,Hz). The end-to-end cue-to-render latency under our live configuration on the RTX~$5090$ deployment hardware (main-paper Sec.~\ref{sec:live_system}) is decomposed per stage in Tab.~\ref{tab:latency_supp}; that breakdown reports the single-emit forward time (model and avatar), whereas Tab.~\ref{tab:fps_supp} reports amortised forward-only throughput on the L40 baseline hardware, so the two numbers measure different quantities.

\begin{table}[h]
\small
\centering
\caption{End-to-end cue-to-motion latency on RTX $5090$, decomposed per stage.}
\label{tab:latency_supp}
\setlength{\tabcolsep}{8pt}
\begin{tabular}{@{}lr@{}}
\toprule
Stage & Latency \\
\midrule
Token-grid quantisation & $80$\,ms \\
Audio-encoder lookahead & $60$\,ms \\
Model forward          & $80$\,ms \\
Gaussian avatar render & $70$\,ms \\
\midrule
\textbf{Total}         & $\mathbf{\sim\!290}$\,\textbf{ms} \\
\bottomrule
\end{tabular}
\end{table}

\paragraph{Lowest-latency live configuration.}
The live deployment of main-paper Sec.~\ref{sec:live_system} sets the commit stride to $s{=}1$: each forward pass commits a single $4$-frame motion token, i.e.\ $160$\,ms of playback at $25$\,fps. The per-emit forward (\emph{Model forward} $+$ \emph{Gaussian avatar render} in Tab.~\ref{tab:latency_supp}) totals $\sim\!150$\,ms, fitting inside this $160$\,ms playback budget; \emph{Token-grid quantisation} and \emph{Audio-encoder lookahead} are one-time pipeline costs incurred once per partner cue. Together they give the $\sim\!290$\,ms cue-to-render latency in Tab.~\ref{tab:latency_supp}, within the $200$--$500$\,ms reaction-lag range observed in real conversational turn-taking~\cite{stivers2009turntaking}.

\section{Per-Subset Comparison Breakdown}
\label{sec:breakdown_supp}

The main-paper Tab.~2 reports the comparison averaged across the held-out test split ($n{=}765$). Tab.~\ref{tab:sota_speak_supp}, \ref{tab:sota_listen_supp}, and \ref{tab:sota_mixed_supp} break the same comparison down by conversational role: \textsc{Speak} ($n{=}170$, purely speaking), \textsc{Listen} ($n{=}330$, purely listening), and \textsc{Mixed} ($n{=}265$, turn-taking transitions). LipSync\,(mm) is the per-frame mean lip-vertex error; rPCC\textsubscript{f\,v} is reported as ${\times}10^{-4}$. Best per column in bold; cells color-shaded by per-column rank.

\begin{table}[h]
\footnotesize
\centering
\caption{Comparison on the \textsc{Speak} subset ($n{=}170$, target purely speaking).}
\label{tab:sota_speak_supp}
\setlength{\tabcolsep}{3pt}
\resizebox{\linewidth}{!}{%
\begin{tabular}{@{}lccccccccc@{}}
\toprule
{Method} & {LipSync$\downarrow$} & {vFD\textsubscript{E}$\downarrow$} & {FDD\textsubscript{u}$\downarrow$} & {JDD$\downarrow$} & {PDD\textsubscript{h}$\downarrow$} & {Div\textsubscript{E}$\uparrow$} & {Div\textsubscript{H}$\uparrow$} & {RPCC\textsubscript{n}$\downarrow$} & {rPCC\textsubscript{f\,v}$\downarrow$} \\
\midrule
DiffPoseTalk        & 12.07 & 0.0414 & 0.0428 & 0.1635 & 0.4119 & 1.584 & 0.852 & 0.558 & 9.93 \\
DualTalk-FT         & \cellcolor{rank1}\textbf{8.28} & 0.0593 & 0.0680 & 0.2080 & 0.3938 & 1.871 & 0.988 & 0.383 & 6.22 \\
UniLS-FT            & 10.63 & 0.0437 & 0.0483 & 0.1609 & 0.3622 & 1.801 & 1.199 & 0.414 & 6.40 \\
Ours-Distill        & \cellcolor{rank2}8.47 & \cellcolor{rank2}0.0299 & \cellcolor{rank2}0.0287 & \cellcolor{rank1}\textbf{0.1257} & \cellcolor{rank2}0.3203 & \cellcolor{rank2}2.209 & \cellcolor{rank2}1.340 & \cellcolor{rank2}0.363 & \cellcolor{rank1}\textbf{5.32} \\
\textbf{Ours (Full)} & 8.48 & \cellcolor{rank1}\textbf{0.0296} & \cellcolor{rank1}\textbf{0.0279} & \cellcolor{rank2}0.1261 & \cellcolor{rank1}\textbf{0.3085} & \cellcolor{rank1}\textbf{2.214} & \cellcolor{rank1}\textbf{1.401} & \cellcolor{rank1}\textbf{0.360} & \cellcolor{rank2}5.34 \\
\bottomrule
\end{tabular}}
\end{table}

On \textsc{Speak}, \textbf{Ours (Full)} wins six columns (vFD\textsubscript{E}, FDD\textsubscript{u}, PDD\textsubscript{h}, Div\textsubscript{E}, Div\textsubscript{H}, RPCC\textsubscript{n}); \textbf{Ours-Distill} wins two (JDD, rPCC\textsubscript{f\,v}) by margins under $1\%$; DualTalk-FT wins LipSync ($8.28$ vs.\ Ours $8.48$, a $0.2$\,mm gap that we read as a small lip-event-timing edge of the autoregressive sampler).

\begin{table}[h]
\footnotesize
\centering
\caption{Comparison on the \textsc{Listen} subset ($n{=}330$, target purely listening).}
\label{tab:sota_listen_supp}
\setlength{\tabcolsep}{3pt}
\resizebox{\linewidth}{!}{%
\begin{tabular}{@{}lccccccccc@{}}
\toprule
{Method} & {LipSync$\downarrow$} & {vFD\textsubscript{E}$\downarrow$} & {FDD\textsubscript{u}$\downarrow$} & {JDD$\downarrow$} & {PDD\textsubscript{h}$\downarrow$} & {Div\textsubscript{E}$\uparrow$} & {Div\textsubscript{H}$\uparrow$} & {RPCC\textsubscript{n}$\downarrow$} & {rPCC\textsubscript{f\,v}$\downarrow$} \\
\midrule
DiffPoseTalk        & 12.11 & 0.0342 & 0.0388 & 0.1183 & 0.2121 & 0.930 & 0.604 & 0.448 & 8.44 \\
DualTalk-FT         & 9.34 & 0.0554 & 0.0627 & 0.1947 & 0.2797 & 0.300 & 0.253 & 0.416 & 5.85 \\
UniLS-FT            & 11.30 & 0.0374 & 0.0430 & 0.1221 & 0.2609 & 0.677 & 0.647 & 0.514 & 7.64 \\
Ours-Distill        & \cellcolor{rank2}8.83 & \cellcolor{rank2}0.0250 & \cellcolor{rank2}0.0251 & \cellcolor{rank2}0.0935 & \cellcolor{rank2}0.1834 & \cellcolor{rank2}1.087 & \cellcolor{rank2}0.686 & \cellcolor{rank2}0.412 & \cellcolor{rank1}\textbf{4.94} \\
\textbf{Ours (Full)} & \cellcolor{rank1}\textbf{8.79} & \cellcolor{rank1}\textbf{0.0245} & \cellcolor{rank1}\textbf{0.0245} & \cellcolor{rank1}\textbf{0.0903} & \cellcolor{rank1}\textbf{0.1781} & \cellcolor{rank1}\textbf{1.133} & \cellcolor{rank1}\textbf{0.822} & \cellcolor{rank1}\textbf{0.394} & \cellcolor{rank2}5.04 \\
\bottomrule
\end{tabular}}
\end{table}

On \textsc{Listen}, \textbf{Ours (Full)} wins eight of nine columns, with Ours-Distill winning rPCC\textsubscript{f\,v} by $2\%$ ($4.94$ vs.\ $5.04$). The gap on \textsc{Listen} Div\textsubscript{H} between Ours ($0.822$) and DualTalk-FT ($0.253$) is the largest single-column gap in the comparison.

\begin{table}[h]
\footnotesize
\centering
\caption{Comparison on the \textsc{Mixed} subset ($n{=}265$, the target alternates between speaking and listening, exercising turn-taking transitions).}
\label{tab:sota_mixed_supp}
\setlength{\tabcolsep}{3pt}
\resizebox{\linewidth}{!}{%
\begin{tabular}{@{}lccccccccc@{}}
\toprule
{Method} & {LipSync$\downarrow$} & {vFD\textsubscript{E}$\downarrow$} & {FDD\textsubscript{u}$\downarrow$} & {JDD$\downarrow$} & {PDD\textsubscript{h}$\downarrow$} & {Div\textsubscript{E}$\uparrow$} & {Div\textsubscript{H}$\uparrow$} & {RPCC\textsubscript{n}$\downarrow$} & {rPCC\textsubscript{f\,v}$\downarrow$} \\
\midrule
DiffPoseTalk        & 13.14 & 0.0395 & 0.0410 & 0.1562 & 0.3459 & 1.423 & 0.788 & 0.475 & 8.69 \\
DualTalk-FT         & \cellcolor{rank1}\textbf{9.98} & 0.0581 & 0.0654 & 0.2046 & 0.3513 & 1.343 & 0.838 & \cellcolor{rank1}\textbf{0.381} & \cellcolor{rank1}\textbf{6.39} \\
UniLS-FT            & 12.79 & 0.0391 & 0.0436 & 0.1365 & 0.3574 & 1.301 & 1.090 & \cellcolor{rank2}0.415 & 7.69 \\
Ours-Distill        & 10.06 & \cellcolor{rank2}0.0267 & \cellcolor{rank2}0.0275 & \cellcolor{rank2}0.0992 & \cellcolor{rank2}0.2676 & \cellcolor{rank2}1.737 & \cellcolor{rank2}1.116 & 0.463 & 6.65 \\
\textbf{Ours (Full)} & \cellcolor{rank2}10.04 & \cellcolor{rank1}\textbf{0.0261} & \cellcolor{rank1}\textbf{0.0267} & \cellcolor{rank1}\textbf{0.0975} & \cellcolor{rank1}\textbf{0.2566} & \cellcolor{rank1}\textbf{1.774} & \cellcolor{rank1}\textbf{1.214} & 0.451 & \cellcolor{rank2}6.61 \\
\bottomrule
\end{tabular}}
\end{table}

\textsc{Mixed} is the only subset on which DualTalk-FT beats Ours on partner-coupling: it wins RPCC\textsubscript{n} ($0.381$ vs.\ Ours $0.451$) and rPCC\textsubscript{f\,v} ($6.39$ vs.\ Ours $6.61$). DualTalk-FT also wins LipSync by $0.06$\,mm ($9.98$ vs.\ Ours $10.04$, the same tied range observed on \textsc{Speak}). Ours (Full) wins the remaining six fidelity and diversity columns, so the partner-coupling gap on \textsc{Mixed} is specific to turn-taking transitions rather than a general partner-coupling weakness.

\section{Training-Objective Ablation}

We ablate the velocity-energy-floor losses (\emph{$-$\,no energy}; main-paper Sec.~3.4) and the style-conditioning pathway (\emph{$-$\,no style}; main-paper Sec.~3.3) and report results in Tab.~\ref{tab:ablations_supp}. The metric set and conventions match main-paper Tab.~2 (best per column in bold; cells color-shaded by per-column rank).

\begin{table*}[!t]
\footnotesize
\centering
\caption{Training-objective ablation, split by conversational role: \textsc{Speak} ($n{=}170$, purely speaking) and \textsc{Listen} ($n{=}330$, purely listening). LipSync on \textsc{Speak} is the vertex-space mouth-region L2 in mm; rPCC\textsubscript{f\,v} reported as ${\times}10^{-4}$. Best per column in bold; cells color-shaded by per-column rank.}
\label{tab:ablations_supp}
\setlength{\tabcolsep}{2.5pt}
\begin{tabular}{@{}lcccccccc!{\quad}ccccccc@{}}
\toprule
& \multicolumn{8}{c!{\quad}}{\textsc{Speak} ($n{=}170$)} & \multicolumn{7}{c}{\textsc{Listen} ($n{=}330$)} \\
\cmidrule(lr){2-9}\cmidrule(lr){10-16}
{Variant} & {LipSync\,(mm)$\downarrow$} & {vFD\textsubscript{E}$\downarrow$} & {FDD\textsubscript{u}$\downarrow$} & {JDD$\downarrow$} & {PDD\textsubscript{h}$\downarrow$} & {Div\textsubscript{E}$\uparrow$} & {Div\textsubscript{H}$\uparrow$} & {rPCC\textsubscript{f\,v}$\downarrow$} & {vFD\textsubscript{E}$\downarrow$} & {FDD\textsubscript{u}$\downarrow$} & {PDD\textsubscript{h}$\downarrow$} & {Div\textsubscript{E}$\uparrow$} & {Div\textsubscript{H}$\uparrow$} & {RPCC\textsubscript{n}$\downarrow$} & {rPCC\textsubscript{f\,v}$\downarrow$} \\
\midrule
\textbf{Ours (Full)} & \cellcolor{rank1}\textbf{8.48} & \cellcolor{rank1}\textbf{0.0296} & \cellcolor{rank1}\textbf{0.0279} & \cellcolor{rank1}\textbf{0.126} & \cellcolor{rank2}0.309 & \cellcolor{rank1}\textbf{2.214} & \cellcolor{rank2}1.401 & \cellcolor{rank2}5.34 & \cellcolor{rank1}\textbf{0.0245} & \cellcolor{rank1}\textbf{0.0245} & \cellcolor{rank1}\textbf{0.178} & \cellcolor{rank2}1.133 & \cellcolor{rank1}\textbf{0.822} & \cellcolor{rank2}0.391 & \cellcolor{rank2}5.04 \\
$-$ no energy       & 9.18 & 0.0346 & 0.0390 & \cellcolor{rank2}0.126 & 0.332 & \cellcolor{rank2}2.163 & 1.362 & 5.47 & 0.0342 & 0.0391 & 0.218 & 1.056 & 0.701 & \cellcolor{rank1}\textbf{0.387} & 5.20 \\
$-$ no style        & \cellcolor{rank2}8.87 & \cellcolor{rank2}0.0318 & \cellcolor{rank2}0.0289 & 0.130 & \cellcolor{rank1}\textbf{0.291} & 2.143 & \cellcolor{rank1}\textbf{1.516} & \cellcolor{rank1}\textbf{5.24} & \cellcolor{rank2}0.0252 & \cellcolor{rank2}0.0250 & \cellcolor{rank2}0.186 & \cellcolor{rank1}\textbf{1.144} & \cellcolor{rank1}\textbf{0.822} & 0.404 & \cellcolor{rank1}\textbf{4.97} \\
\bottomrule
\end{tabular}
\end{table*}

Removing the energy-floor losses ($-$\,\emph{no energy}) is the most damaging variant. On \textsc{Listen}, FDD\textsubscript{u} increases by $60\%$, vFD\textsubscript{E} by $40\%$, PDD\textsubscript{h} by $22\%$, and head-pose diversity decreases by $15\%$ (Div\textsubscript{H} $0.822 \to 0.701$). On \textsc{Speak} the same pattern produces the largest LipSync regression in our ablation ($+8\%$, $9.18$\,mm) together with vFD\textsubscript{E} $+17\%$ and FDD\textsubscript{u} $+40\%$. The energy floors are therefore central to upper-face and lip structure on both regimes.

Removing the style-conditioning pathway ($-$\,\emph{no style}) costs lip sync and dynamics tightness on \textsc{Speak} (LipSync $+4.6\%$ at $8.87$\,mm, vFD\textsubscript{E} $+7\%$, JDD $+3\%$) but is roughly neutral on \textsc{Listen} --- $-$\,\emph{no style} stays within $3\%$ of \emph{Full} on the listening fidelity columns. Head-pose diversity on \textsc{Speak} actually rises ($1.401 \to 1.516$, $+8\%$): without the style channel the speaking-side head motion becomes less style-locked and gains entropy. The motion-style code is therefore most useful on the speaking side, where it ties head and face dynamics to the target's motion idiosyncrasy.

\FloatBarrier

\section{Partner-Conditioning Ablation Tables}
\label{sec:partner_ablation_supp}

The main-paper Sec.~\ref{sec:ablations} discusses the listening-side partner-conditioning ablation; Tab.~\ref{tab:partner_modality} reports the supporting numbers.

\begin{table}[H]
\centering
\caption{Partner-conditioning ablation on the \textsc{Listen} subset ($n{=}330$). rPCC\textsubscript{f\,v} reported as ${\times}10^{-4}$. Best per column in bold; cells color-shaded by per-column rank.}\label{tab:partner_modality}
\setlength{\tabcolsep}{3pt}
\footnotesize
\resizebox{\linewidth}{!}{%
\begin{tabular}{@{}lccccccc@{}}
\toprule
{Variant} & {vFD\textsubscript{E}$\downarrow$} & {FDD\textsubscript{u}$\downarrow$} & {PDD\textsubscript{h}$\downarrow$} & {Div\textsubscript{E}$\uparrow$} & {Div\textsubscript{H}$\uparrow$} & {RPCC\textsubscript{n}$\downarrow$} & {rPCC\textsubscript{f\,v}$\downarrow$} \\
\midrule
\textbf{Full}                          & \cellcolor{rank1}\textbf{0.0245} & \cellcolor{rank2}0.0245 & \cellcolor{rank1}\textbf{0.178} & \cellcolor{rank1}\textbf{1.133} & \cellcolor{rank2}0.822 & \cellcolor{rank1}\textbf{0.391} & \cellcolor{rank1}\textbf{5.04} \\
$-$ partner audio                      & \cellcolor{rank2}0.0248 & \cellcolor{rank1}\textbf{0.0244} & \cellcolor{rank2}0.180 & \cellcolor{rank2}1.131 & \cellcolor{rank1}\textbf{0.901} & \cellcolor{rank2}0.403 & \cellcolor{rank2}5.46 \\
$-$ partner motion                     & 0.0251 & 0.0253 & 0.190 & 0.826 & 0.442 & 0.687 & 6.30 \\
$-$ partner stream                     & 0.0257 & 0.0265 & 0.195 & 0.784 & 0.424 & 0.670 & 6.54 \\
\bottomrule
\end{tabular}}
\end{table}

We additionally report the same ablation on the \textsc{Speak} subset ($n{=}170$).

\begin{table}[H]
\footnotesize
\centering
\caption{Partner-conditioning ablation on the speaking subset ($n{=}170$). JDD reported $\times 10^{-3}$. Best per column in bold; cells color-shaded by per-column rank.}
\label{tab:partner_modality_speak_supp}
\setlength{\tabcolsep}{3pt}
\resizebox{\linewidth}{!}{%
\begin{tabular}{@{}lccccccc@{}}
\toprule
{Variant} & {LipF1$\uparrow$} & {vFD\textsubscript{E}$\downarrow$} & {FDD\textsubscript{u}$\downarrow$} & {JDD$\downarrow$} & {PDD\textsubscript{h}$\downarrow$} & {Div\textsubscript{E}$\uparrow$} & {Div\textsubscript{H}$\uparrow$} \\
\midrule
\textbf{Full}                          & \cellcolor{rank1}\textbf{0.454} & 0.032 & \cellcolor{rank1}\textbf{0.026} & \cellcolor{rank1}\textbf{1.8} & \cellcolor{rank1}\textbf{0.294} & \cellcolor{rank1}\textbf{2.321} & \cellcolor{rank1}\textbf{1.540} \\
$-$ partner audio                      & \cellcolor{rank2}0.450 & 0.032 & \cellcolor{rank1}\textbf{0.026} & \cellcolor{rank1}\textbf{1.8} & \cellcolor{rank1}\textbf{0.294} & \cellcolor{rank2}2.310 & \cellcolor{rank2}1.534 \\
$-$ partner motion                     & 0.436 & \cellcolor{rank1}\textbf{0.031} & \cellcolor{rank2}0.029 & \cellcolor{rank2}1.9 & 0.314 & 1.986 & 0.990 \\
$-$ partner stream                     & 0.421 & \cellcolor{rank1}\textbf{0.031} & 0.031 & \cellcolor{rank2}1.9 & \cellcolor{rank2}0.313 & 1.937 & 0.983 \\
\bottomrule
\end{tabular}}
\end{table}

\rev{%
\paragraph{Smile-timing F1.}
To quantify the mirror-laugh evidence of main-paper Fig.~\ref{fig:mirror_laugh} at event level, we compute a smile-timing F1 inspired by AV-Flow~\cite{chatziagapi2025avflow}: per clip, a frame counts as smiling when the FLAME mouth-corner distance exceeds the ground-truth clip's $70$th percentile, and we report the frame-level F1 between predicted and ground-truth smile flags, a geometric score with no classifier in the loop. Over the full test pool, STEER reproduces the timing of the listener's smiles best (F1 $0.37$ vs.\ UniLS $0.32$, DualTalk $0.29$, DiffPoseTalk $0.20$). Zeroing partner motion at inference lowers the \textsc{Listen}-subset score by $21\%$, while zeroing partner audio changes it only marginally: the mirrored smile is driven predominantly by the partner's facial motion. We chose smiles as the listener response most reliably mirrored between partners.
}%

LipF1 decreases monotonically along Full $\to$ $-$\,partner audio $\to$ $-$\,partner motion $\to$ $-$\,partner stream ($0.454 \to 0.450 \to 0.436 \to 0.421$), indicating that on the speaking subset partner inputs continue to contribute to lip-event timing fidelity. Div\textsubscript{E} and Div\textsubscript{H} drop by $14\%$ and $36\%$ respectively when partner motion is removed, indicating that without partner motion the model produces flatter, less varied facial and head dynamics.

The lower vFD\textsubscript{E} of the partner-removed variants ($0.031$ vs.\ Full $0.032$) is not an improvement but a signature of a flatter mouth, as evidenced by the simultaneous drop in Div\textsubscript{E} and Div\textsubscript{H}.

\FloatBarrier

\end{document}